\documentclass[acmtog]{acmart}

\usepackage{booktabs} 
\usepackage{graphicx}
\usepackage{amsmath}
\usepackage{booktabs}
\usepackage{float}
\usepackage{multirow}
\usepackage{ulem}
\usepackage{stmaryrd}
\usepackage{siunitx}
\usepackage{textcomp}
\usepackage{titlesec}

\usepackage{bm}

\newcommand{\PAR}[1]{\noindent{\bf #1}}

\usepackage{color}
\definecolor{url_color}{RGB}{42, 83, 163}

\DeclareMathOperator{\similarity}{sim}
\DeclareCaptionFormat{cont}{#1 (cont.)#2#3\par}

\makeatletter
\DeclareRobustCommand\onedot{\futurelet\@let@token\@onedot}
\def\@onedot{\ifx\@let@token.\else.\null\fi\xspace}

\def\eg{\emph{e.g}\onedot} 
\def\ie{\emph{i.e}\onedot}

\def\wrt{w.r.t\onedot} 
\def\etal{\emph{et al}\onedot}
\makeatother

\usepackage[ruled]{algorithm2e} 

\SetAlFnt{\small}
\SetAlCapFnt{\small}
\SetAlCapNameFnt{\small}
\SetAlCapHSkip{0pt}

\setcopyright{rightsretained}
\acmJournal{TOG}
\acmYear{2022}\acmVolume{41}\acmNumber{4}\acmArticle{101}\acmMonth{7}\acmDOI{10.1145/3528223.3530163}

\begin{document}

\title{Neural Rendering in a Room: Amodal 3D Understanding and Free-Viewpoint Rendering for the Closed Scene Composed of\\ Pre-Captured Objects
}

\author{Bangbang Yang}
\email{ybbbbt@gmail.com}
\affiliation{
  \institution{State Key Lab of CAD\&CG, Zhejiang University}
  \country{China}
}

\author{Yinda Zhang}
\email{yindaz@gmail.com}
\affiliation{
  \institution{Google}
  \country{USA}
}

\author{Yijin Li}
\email{eugenelee@zju.edu.cn}
\affiliation{
  \institution{State Key Lab of CAD\&CG, Zhejiang University}
  \country{China}
}

\author{Zhaopeng Cui}
\authornote{Guofeng Zhang and Zhaopeng Cui are corresponding authors.}
\email{zhpcui@gmail.com}
\affiliation{
  \institution{State Key Lab of CAD\&CG, Zhejiang University}
  \country{China}
}

\author{Sean Fanello}
\email{seanfa@google.com}
\affiliation{
  \institution{Google}
  \country{USA}
}

\author{Hujun Bao}
\email{baohujun@zju.edu.cn}
\affiliation{
  \institution{State Key Lab of CAD\&CG, Zhejiang University}
  \country{China}
}

\author{Guofeng Zhang}
\email{zhangguofeng@zju.edu.cn}
\authornotemark[1]
\affiliation{
  \institution{State Key Lab of CAD\&CG, Zhejiang University}
  \country{China}
}

\begin{teaserfigure}
  \centering
    \includegraphics[width=1.0\linewidth]{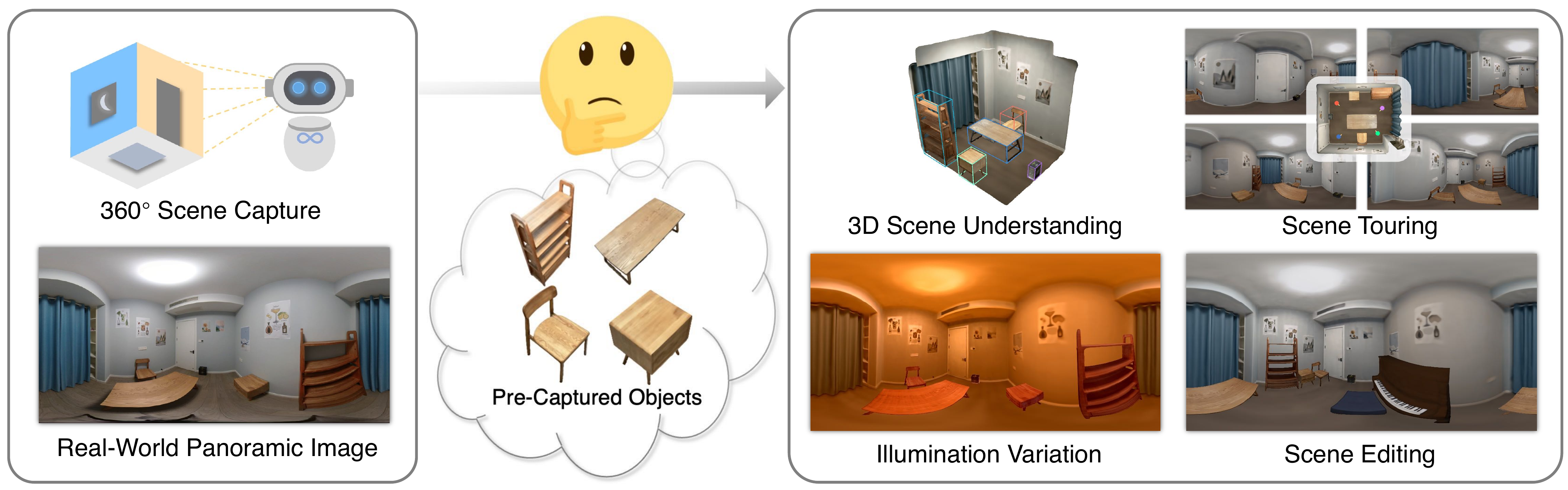}
    \caption{
    \textbf{Motivation.} We focus on a scenario where a service robot operates in a specific indoor environment (\eg, household, office, or museum). Therefore, it can collect information of the closed scene in an offline stage, 
    then provide effective amodal scene understanding with a single panoramic capture of the room, which facilitates high-level tasks and delivers immersive synchronized free-viewpoint touring with illumination variation and scene editing.
}
    \label{fig:teaser}
\end{teaserfigure}

\begin{abstract}
\noindent We, as human beings, can understand and picture a familiar scene from arbitrary viewpoints given a single image, whereas this is still a grand challenge for computers.
We hereby present a novel solution to mimic such human perception capability based on a new paradigm of amodal 3D scene understanding with neural rendering for a closed scene.
Specifically, we first learn the prior knowledge of the objects in a closed scene via an offline stage, which facilitates an online stage to understand the room with unseen furniture arrangement.
During the online stage, given a panoramic image of the scene in different layouts, we utilize a holistic neural-rendering-based optimization framework to efficiently estimate the correct 3D scene layout and deliver realistic free-viewpoint rendering. 
In order to handle the domain gap between the offline and online stage, our method exploits compositional neural rendering techniques for data augmentation in the offline training.
The experiments on both synthetic and real datasets demonstrate that our two-stage design achieves robust 3D scene understanding and outperforms competing methods by a large margin, and we also show that our realistic free-viewpoint rendering enables various applications, including scene touring and editing.
Code and data are available on the project webpage:
\urlstyle{tt}
\textcolor{url_color}{\url{https://zju3dv.github.io/nr_in_a_room/}}.
\end{abstract}

%
%


%
%


\maketitle

\section{Introduction}
\noindent Given a photo of our living room, as human beings, we can vividly picture the whole layout in our mind, including how the furniture is placed in 3D space and how the environment looks from any viewpoint, even when objects are re-arranged differently in the room.
Granting the computer similar skills would require reliable indoor scene 3D semantic understanding and free-viewpoint rendering capabilities, which ideally are all fulfilled from widely available input, \eg, a single photo.
Over decades, enormous efforts have been made in the field of computer vision and graphics~\cite{total3d, deeppanocontext, light_field, lumigraph, nerf, point_cloud_rendering}, yet the gap with the human perception is still huge.
Despite this, we argue that likely humans are better at this task for places they are familiar with, and the learned prior knowledge on the objects and their arrangement in a closed room are the key to the success.

In this paper, we present a novel
solution for reliable 3D indoor scene understanding and free-viewpoint rendering in a closed scene -- a.k.a. a room with a fixed set of pre-captured objects but placed under unknown arrangements and diverse illuminations.
Inspired by human amodal perception, our method takes advantage of an offline stage to collect prior knowledge of the target scene, where models for each object, \eg, for localization or neural rendering, can be built with an affordable workload and then fine-tuned in the specific scenario for better performance.
With the help of this strong prior knowledge, during the online stage, our method only needs light-weighted input, \ie, a single panoramic image taken from the scene, and can reliably recognize and localize objects in 3D space and render the scene from arbitrary camera viewpoints via amodal 3D understanding.
While general scene understanding~\cite{total3d,deeppanocontext} makes the best effort to make predictions under unseen environments but still suffers from generalization issues, our amodal scene understanding aims at an accurate and reliable scene understanding for familiar scenes.

A flexible yet effective scene representation is critical to the considered task.
Traditional representations such as textured meshes~\cite{Waechter2014Texturing,liu2019soft,im2cad} or voxels~\cite{kim20133d,song2017semantic} generally have some drawbacks, \eg, limited rendering quality~\cite{liu2019soft} and resolution~\cite{song2017semantic}, requiring pre-built CAD furniture model for scene reconstruction~\cite{im2cad} and explicit lighting/material definitions for lighting variations~\cite{matusik2003data,li2020inverse}, which prohibits fine-grained scene rendering and understanding. 
We thus choose the neural implicit representation~\cite{nerf} as it enables geometric reconstruction with photo-realistic volumetric rendering, and it could be extended to support functionalities such as appearance variation~\cite{nerf_w} and scene graph decomposition~\cite{neural_scene_graph} with rendering-based optimization.

Specifically, we first build object detection and 3D pose estimation models for all the objects of interest as well as a neural rendering model for each object, including the empty room.
At run-time, given a panoramic image taken from the room stuffed with pre-captured objects in a new arrangement, the scene understanding task can be achieved by 3D object detection and pose estimation, followed by an optimization via differentiable rendering using the neural rendering models.
Additionally, the per-object neural rendering models can be plugged in to support full scene free-viewpoint rendering.
While this framework is technically plausible, we find it suffers from several challenges as follows, which we will address in this work:

\noindent\textbf{Intensive Computation.}
Neural volume rendering methods are typically computationally intensive since a tremendous number of network queries are required for points densely sampling along pixel rays, making it prohibitive for back-propagation-based optimization, like pose estimation, where the rendering needs to be done repetitively.
iNeRF~\cite{inerf} mitigates this issue by restricting sample pixels inside the detected region of interest, which reduces the computation cost and enables the camera pose estimation with respect to a single object on a commodity-level GPU. However, this is still not practical for room-scale scenarios when multiple objects need to be jointly optimized in order to handle mutual occlusions or physical relations.
To tackle this challenge, we learn an implicit surface model jointly with its radiance field, inspired by NeuS~\cite{neus}, which allows us to perform efficient sphere tracing~\cite{dist} at the early stage of the rendering, leveraging the estimated ray-to-surface distances.
Points can then be sampled from regions close to the surface, and a small number of points is sufficient for the optimization.
In this way, we significantly reduce the computational cost and make it feasible to finish the joint optimization with multiple objects on a single GPU in a reasonable amount of computation.

\noindent\textbf{Incorrect Physical Relationship.}
Even though machine learning models are trained per-scene, they could still make obvious mistakes like breaking the physical rules and resulting in implausible novel view rendering, \eg, objects flying in the air or intersecting with walls.
To solve this problem, we propose several novel physical losses and integrate a physics-based optimization into the neural-rendering-based optimization, where the conformity to prior knowledge and even pre-defined rules (\eg, a bed should attach to the wall) are jointly optimized with the photometric error between the rendered image and the observation.
This significantly helps fix errors made on individual objects and improves the overall object pose accuracy, which further delivers context abides rendering.

\begin{figure*}[t!]
    \centering
    \includegraphics[width=1.0\linewidth, trim={0 0 0 0}, clip]{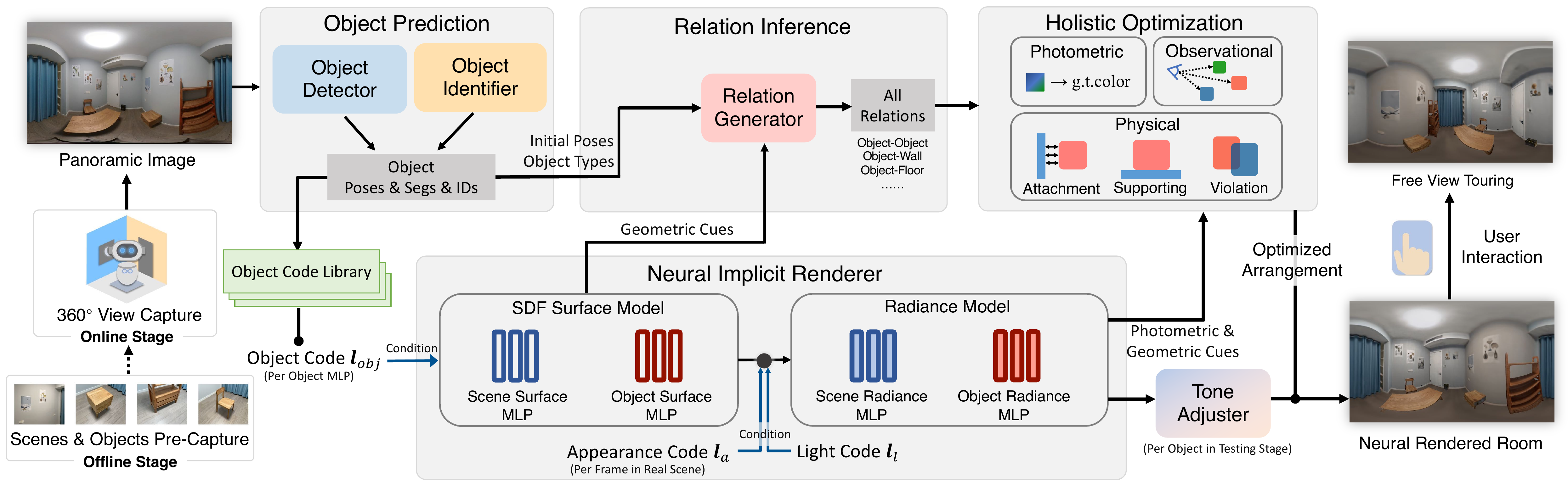}
    \caption{
    During the offline stage, we learn a neural implicit renderer and object detectors with pre-captured scene and objects.
    During the online stage, given a panoramic capture of a room, we first recognize object identities and estimate object meta information.
    Then, we generate object relations based on the prediction and the geometric cues from the renderer.
    Finally, we conduct a holistic optimization to obtain 3D scene understanding by jointly optimizing all the photometric and geometric cues.
    }
    \label{fig:framework}
\end{figure*}

\noindent\textbf{Domain Gap.}
The lighting condition may inevitably vary in the scene, and object renderings from the models trained at offline stage may not be consistent with the environment, which will further influence the rendering-based optimization.
To mitigate this, we propose to exploit compositional neural rendering to augment the training data.
In particular, we augment the pre-captured data with environment maps sourced from polyhaven.com~\cite{HDRIHaven}, and learn the neural rendering models conditioned on lighting represented in a latent space.
During the neural rendering based optimization, the neural rendering model is able to respond to novel illumination other than the one during the pre-capture stage, and both the environment lighting and object pose can be successfully optimized.
We also synthesize objects with different scene layouts and render photo-realistic images for the training of object prediction, which empirically enhances model robustness.

Our contributions can be summarized as follows.
We present a practical solution for a novel task which aims at amodal 3D scene understanding and free-viewpoint rendering for indoor environments from a single panoramic image. 
We design a two-stage framework in which per-object pre-trained models are learned offline, and a neural-rendering-based optimization is exploited for online 3D understanding.
We analyze the technical challenges for this novel task and propose mitigation techniques to improve run-time efficiency, add physical constraints in the model, handle illumination changes, and increase the data diversity that is hard to be ensured physically.
Extensive experiments show that our method can achieve significantly better 3D scene understanding performance than state-of-the-art general 3D scene understanding methods and meanwhile, deliver free-viewpoint rendering capability that supports high-level applications like scene editing and virtual touring.

\section{Related Work}

\noindent\textbf{3D Scene Understanding.}
3D scene understanding is a popular topic in computer vision.
Early works mainly focus on room layout estimation with Manhattan World~\cite{manhattan, manhattan_world, horizonnet,zou2018layoutnet,yan20203d} or cuboid assumption~\cite{delay, mallya2015learning}.
Song~\etal~\cite{robot_in_a_room} attempts to reconstruct and recognize scene objects from a domestic robot but requires laborious crowd-sourced annotations.
With the advance of neural networks, many works propose to estimate both object poses and the room layout~\cite{scene_parsing,unifying_holistic,deepcontext}.
To recover object shapes, some methods~\cite{pixel2mesh, atlasnet,chen2019learning} reconstruct meshes from a template, and others~\cite{im2cad, holistic} adopt shape retrieval approaches to search from a given CAD database.
Recently, some approaches~\cite{total3d,im3d,dahnert2021panoptic,popov2020corenet,yang2016efficient,yang2019dula} enable 3D scene understanding by generating a room layout, camera pose, object bounding boxes, or even meshes from a single view, automatically completing and annotating scene meshes~\cite{bokhovkin2021towards} or predicting object alignments and layouts~\cite{avetisyan2020scenecad} from an RGB-D scan.
Inspired by PanoContext~\cite{panocontext} that panoramic images contain richer context information than the perspective ones, Zhang \etal~\cite{deeppanocontext} propose a better 3D scene understanding method with panoramic captures as input.
For amodal scene completion, Zhan \etal~\cite{zhan2020self} propose to decompose cluttered objects of an image into individual identities.
However, these works still suffer from limited generalization in real-world environments and do not allow fine-grained scene presence from arbitrary views.

\noindent\textbf{Neural Rendering.}
Neural rendering methods aim at synthesizing novel views of objects and scene by learning scene representation from 2D observations in various forms, such as voxels~\cite{deepvoxel, neural_volume}, point clouds~\cite{point_cloud_rendering}, meshes~\cite{fvs, svs}, multi-plane images~\cite{view_mpi, llff,wang2021ibrnet} and implicit functions~\cite{nerf,SRN,niemeyer2020differentiable_implicit3}.
NeRF ~\cite{nerf} uses volume rendering to achieve photo-realistic results; follow up works extend the model to multiple tasks, such as pose estimation~\cite{inerf}, dense surface reconstruction~\cite{neus,volsdf,unisurf} and scene editing~\cite{object_nerf,object_centric_nerf,granskog2021neural}.
Meanwhile, other methods~\cite{fvs, svs} also show impressive free-viewpoint rendering capability in the wild, or scene rendering~\cite{luo2020end,devries2021unconstrained} of indoor environments.
However, existing neural rendering pipelines either need to be trained for a static scene thus do not generalize to dynamic environments, or require domain prior~\cite{wang2021ibrnet,yu2021pixelnerf}, limiting the free-viewpoint rendering in unconstrained settings.

\section{Method}
\noindent 
Given a panoramic image of a closed environment with unknown furniture placement, our goal is to achieve reliable 3D scene understanding, including instance semantic detection, 3D geometry of each object, and their arrangements (\ie, object positions) in the room, utilizing the data pre-captured beforehand.
We split the whole pipeline into an offline stage and an online stage.
During the \textbf{offline stage}, we scan each object and the scene background with an RGB-D camera, train a neural implicit renderer for every object of interest in the room, and then fine-tune object detectors via compositional neural rendering.
In the \textbf{online stage}, as shown in Fig.~\ref{fig:framework}, we first predict object meta information (\ie, poses, segmentation and IDs) from the panoramic image, and then follow pre-defined rules to generate object-object and object-room relations based on the object prediction and geometric cues (\eg, physical distances obtained from the encoded neural implicit model).
Finally, to correctly estimate the scene arrangement and lighting condition that visually fits the input panorama, we perform holistic optimization with all the photometric and geometric cues, which further enables free-viewpoint scene touring and scene editing.

\subsection{Offline Stage}

\subsubsection{Neural Implicit Renderer}
\label{ssec:implicit_neural_renderer}
\hfill\\
\PAR{Neural Implicit Model for Scene and Objects.}
We use a neural implicit renderer that hierarchically encodes the room.
Practically, we choose the SDF-based implicit field for geometry representation~\cite{neus,volsdf}\footnote{In our paper, we use the formulation from NeuS~\cite{neus}, but VolSDF~\cite{volsdf} is also applicable.}, since it provides an exact surface to facilitate geometric optimization, \eg, for collision detection, while NeRF's density field is too noisy or uncertain to support a similar objective.
As shown in Fig.~\ref{fig:framework}, we separately express geometry in SDF values (with SDF surface model $F_{\text{SDF}}$) and colors (with radiance model $F_{\text{R}}$).
We will show later that this formulation enables efficient neural-rendering-based optimization by providing geometric cues like ray intersection distances with sphere tracing.
Motivated by Yang \etal~\cite{object_nerf}, we encode scene background and objects in two branches, and use the object code $\bm{l}_{\text{obj}}$ to control the visibility of a certain object, rather than per-model per-object training.
We render the object $k$ with sampled points $\{\mathbf{x}_i|i=1,...,N\}$ along the ray $\bm{r}$, which is defined as:
\begin{equation}
\begin{split}
    &\hat{C}(\bm{r})_{\text{obj}} = \sum_{i=1}^{N}
    T_i \alpha_i {\mathbf{c}_{\text{obj}}}_i, \;\;
    T_i = \prod_{j=1}^{i-1}(1-{\alpha_{\text{obj}}}_j), \\
    & 
    {\alpha_{\text{obj}}}_j = \max \left (\frac{\Phi_s(\text{SDF}(\mathbf{x}_{i})_j) - \Phi_s(\text{SDF}(\mathbf{x}_{i+1})_j)}{\Phi_s(\text{SDF}(\mathbf{x}_i)_j)}, 0 \right)
    .
\end{split}
\end{equation}
Note that we omit the object index $k$ for brevity.
$T_i$ is the accumulated transmittance, $\Phi_s$ is the logistic density distribution, $\text{SDF}(\mathbf{x})=F_{\text{SDF}}(\mathbf{x},l_{\text{obj}})$, and $\alpha_{\text{obj}}$ is the opacity value derived from the SDF surface model.
$\mathbf{c}_{\text{obj}}$ is the color defined as $\mathbf{c}_{\text{obj}}=F_{\text{R}}(\mathbf{x},\mathbf{v},{\bm{l}_{\text{obj}}},{\bm{l}_{a}},{\bm{l}_{l}})$, where $\mathbf{v}$ is the viewing direction,  $\bm{l}_{a}$ is the appearance code~\cite{nerf_w} that handles per-frame sensing variations (\eg, white balance and auto-exposure on real-world data), $\bm{l}_{l}$ is the light code introduced later.
We supervise the renderer with color, depth and object masks, and jointly render multi-objects and scene background by ordering the distance of samples along the ray directions and render pixels $\hat{C}(\bm{r})$ following the quadrature rules.
More details can be found in the supplementary material.

\begin{figure}[t!]
    \centering
    \scalebox{1.0}[1.0]{
    \includegraphics[width=1.0\linewidth, trim={0 0 0 0}, clip]{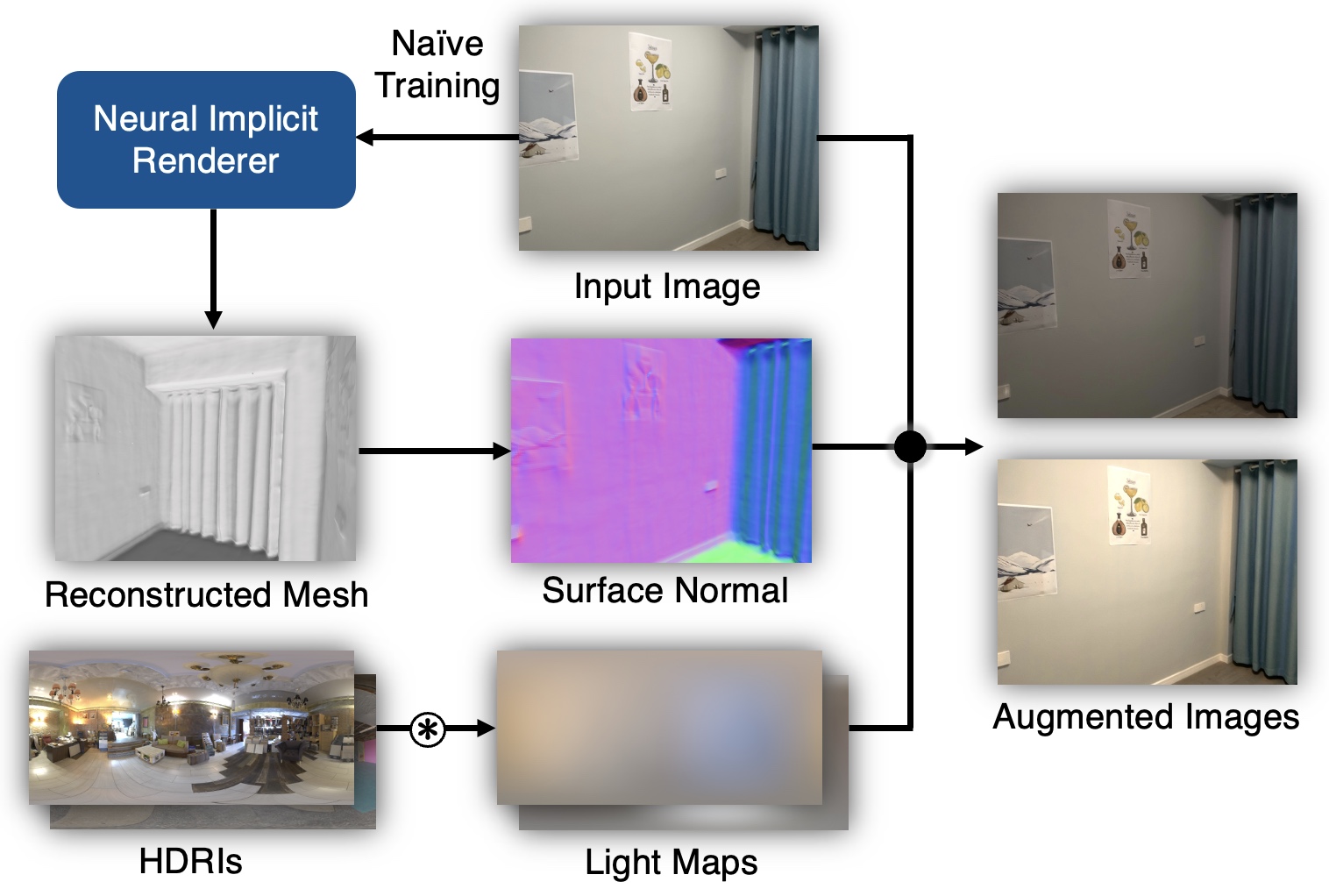}
    }
    \caption{
    \textbf{Lighting augmentation.} We show how to leverage the neural rendering model and pre-convolved HDR maps to synthesize novel lighting conditions. See text for details.
    }
    \label{fig:light_map_aug}
\end{figure}

\PAR{Lighting Augmentation \& Light Code Learning.}
We learn a neural renderer conditioned on a latent lighting space $\bm{l}_{l}$, aiming at modeling scene-level illumination variation and adapting to the target scene depicted in the given panorama.
Since it is non-trivial to capture real-world images with thorough lighting variation, we synthetically augment the captured image with diffuse shading rendered from realistic HDR environment maps.
Practically, we gather 100 HDRI indoor environment maps from the Internet~\cite{HDRIHaven} and convolve them to diffuse irradiance maps~\cite{ibl}. 
Then we compute the per-pixel surface normal in the world coordinate and retrieve the corresponding light intensity from the light map.
Finally, we multiply the light intensity to the input images.
However, for real-world data, reliable surface normals are not readily available.
To tackle this problem, we leverage a two-stage pipeline by first training a na\"ive neural renderer without augmentation, and then extracting mesh from the model for normal computation.
We show this procedure in Fig.~\ref{fig:light_map_aug}, where the captured image has been naturally augmented with two different light maps.
During the training stage, we randomly augment the input images with pre-convolved light maps and feed the radiance model $F_{\text{R}}$ with a learnable light code $\bm{l}_{l}^m$, where $m$ is the index of the light map.
Although such geometry-aware augmentation does not cover every important aspect of the real-world physics and provides augmented data only up to an approximation, it brings convenience to the offline stage: the training data is collected only once under a mild lighting condition, and the network empirically adapts to unseen lighting decently (see Sec.~\ref{ssec:light_ada}).

\subsubsection{3D Object Prediction Fine-tuning}
\hfill\\
\PAR{Module Design.}
As illustrated in Fig.~\ref{fig:framework} (on top left), we adopt the object detector (ODN) from Zhang \etal~\cite{panocontext,deeppanocontext} and Nie \etal~\cite{total3d} to detect scene objects and estimate object poses \wrt the camera, and use an object identifier based on NetVLAD feature similarity~\cite{netvlad} to recognize previously seen objects.

\PAR{Data Augmentation with Neural Rendering.}
At training time for each scene, instead of physically moving objects in the real-world, we exploit compositional neural rendering with implicit neural renderer (Sec.~\ref{ssec:implicit_neural_renderer}) to render labeled panorama for training, where objects are randomly placed following user-defined rules (\eg, bed and table should attach to the floor).
Then, we perform a fast fine-tuning for the pre-trained ODN network from Zhang \etal~\cite{deeppanocontext} and also store the NetVLAD features for each object of different views.

\subsection{Online Stage}

\subsubsection{Bottom-up Initialization}
\label{ssec:init_estim_relation}
\hfill\\
\PAR{Object Prediction.}
We first feed panoramic images to the object detector, and obtain object meta information including initial pose estimation, instance / semantic segmentation and object identities.

\PAR{Relation Generation.}
As demonstrated in ~\cite{total3d,deeppanocontext,panocontext}, indoor scenes are commonly well-structured (\eg, beds and nightstands are usually attached to the wall, desks and chairs are often supported by the floor), and such prior knowledge can be formulated as various relations to enhance arrangement optimization.
Therefore, we also generate a series of relations for physical constraints optimization (Sec.~\ref{ssec:physical_optim}), including object-object support, object-wall attachment and object-floor support.
Practically, we directly infer relations based on object meta information and geometric cues (extracted bounding boxes, ray intersection distance and normal) from the SDF surface model with user-defined rules (see supplementary material).
In theory, our method can also work with rules or scene context learned in a data-driven way~\cite{deeppanocontext}, which we leave for future work.

\PAR{Camera Pose Estimation.}
Since the optimization is based on a known neural implicit model, we need to locate camera poses to ensure background rendering is aligned with the input image. 
To do so, we transform the panorama to multiple perspective views (\ie, similar to "equirectangular to cubemap" conversion by warping pixels according to ray directions) and employ the method from Sarlin \etal~\cite{sarlin2019coarse,sarlin2020superglue} for visual localization.

\PAR{Object Pose Parameterization.}
We optimize poses $\hat{\mathbf{T}}^k \in \text{SE}(3)$ for $K$ objects, where the rotation $\hat{\mathbf{R}}^k$ is parameterized as Zhou \etal~\cite{zhou2019continuity}, and the position (a.k.a. object center) $\hat{\mathbf{p}}^k$ is directly expressed in Euclidean space.

\subsubsection{Photometric Constraint Optimization}
\label{ssec:photo_optim}
\hfill\\
\PAR{Tone Adjuster.}
To better adapt the lighting condition to the input panorama at the online stage, we introduce a per-object tone adjuster which explicitly models lighting variations and helps to reduce the burden of light code optimization.
In practice, we additionally optimize a learnable shifting factor $\mathbf{t}_{\text{obj}}^k$ and scaling factor $\mathbf{s}_{\text{obj}}^k$ for each object $j$ as: $\tilde{\mathbf{c}}_{\text{obj}}^k = (\hat{\mathbf{c}}_{\text{obj}}^k - \mathbf{t}_{\text{obj}})^k / \mathbf{s}_{\text{obj}}^k$,
which can be regarded as color transformation~\cite{reinhard2001color} but in a per-object manner.
We find this explicit representation benefits the lighting adaptation, as demonstrated in our experiments.

\PAR{Photometric Loss with Joint Rendering.}
We use photometric constraint by leveraging joint rendering where the photometric loss is back-propagated to optimize per-object poses and light parameters.
For each input image, we sample $N$ rays on the  object masks and $0.2N$ rays on the background so as to ensure the convergence of both objects and background.
The photometric loss is defined as the squared distance between rendered colors $\hat{C}(\bm{r})$ and pixel colors $C(\bm{r})$ from the input panorama for all the sampled rays $\bm{r} \in N_r$:
\begin{equation}
    L_{pho} =  \frac{1}{|N_r|}\sum_{\bm{r}\in N_r} ||\hat{C}(\bm{r})-C(\bm{r})||^2_2.
\end{equation}

\PAR{Safe-Region Volume Rendering.}
However, neural volume rendering requires hundreds of network queries for each ray, which restricts tasks like pose estimation~\cite{inerf} by only sampling a small bunch of rays due to the limitation of GPU memory.
This is particularly true in our task as one ray might go through 2 or 3 objects at a time when object to object occlusions happen, which results in 2 or 3 times more queries than a single object case.
Fortunately, as our renderer learns an SDF-based representation of the geometry, we can easily determine ray intersections using sphere tracing at the early stage of the rendering.
Inspired by Liu \etal~\cite{dist}, we propose a safe-region volume rendering by first computing ray-to-surface distances with efficient sphere tracing and then sampling much fewer points near the surface for differentiable volume rendering.
Our experiments demonstrate that this strategy significantly reduces network query times and allows us to jointly optimize more objects in cluttered scenes.
Please refer to the supplementary material for more details.

\subsubsection{Observation Constraint Optimization}
\label{ssec:obs_optim}
\hfill\\
\PAR{Observation Loss.}
The initial poses from object prediction may be inaccurate on the dimension of camera-to-object distance due to scale ambiguity, but the observing angles (a.k.a. object center re-projection) on the panoramic view are usually reliable.
Thus, we also add an observation constraint by encouraging closer observing angles of objects between initial pose estimation and the optimized pose, as:
\begin{equation}
    L_{obs} = \sum_{k=1}^{K}||1 - \similarity(\mathbf{p}_{\text{init}}^k - \mathbf{p}_{\text{cam}}, \hat{\mathbf{p}}^k - \mathbf{p}_{\text{cam}})||^{2},
\end{equation}
where $\similarity(\cdot)$ denotes cosine similarity, and $\mathbf{p}_{\text{cam}}$ is the camera center estimated in Sec.~\ref{ssec:init_estim_relation}.

\subsubsection{Physical Constraint Optimization}
\label{ssec:physical_optim}
\hfill\\
Prior scene understanding works~\cite{total3d,im3d,deeppanocontext} mainly build physical constraints upon object bounding boxes and room layout under Manhattan assumption.
Thanks to the precise geometries encoded in the neural SDF model, we can define physical constraints to optimize physical conformity at a finer-grained level.

\begin{figure}[t!]
    \centering
    \includegraphics[width=1.0\linewidth, trim={0 0 0 0}, clip]{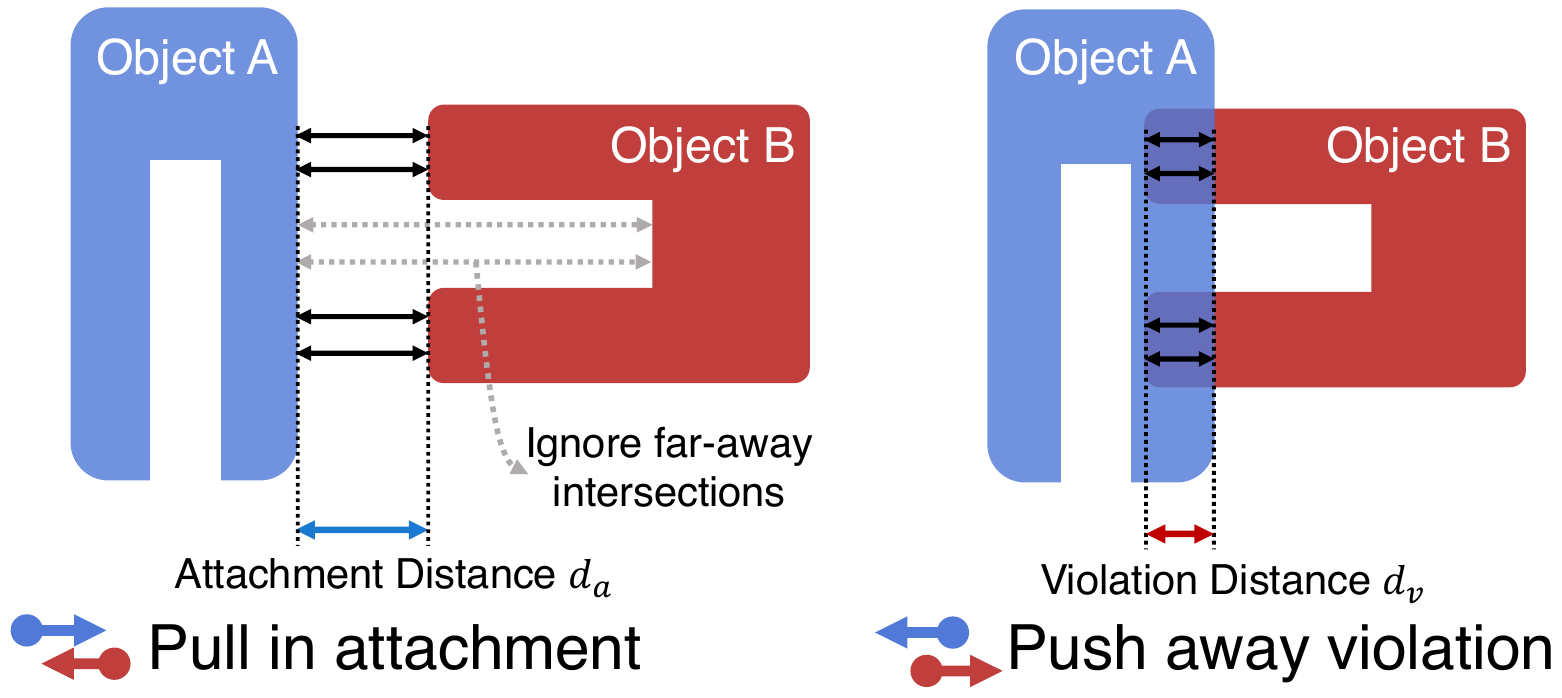}
    \caption{
    \textbf{Magnetic Loss.}
    For the attachment relation, the loss pulls in when two instances are far from each other and pushes away when the violation happens.
    }
    \label{fig:magnetic_loss}
\end{figure}

\PAR{Magnetic Loss.}
We introduce a novel magnetic loss that fully leverages neural renderer's SDF model to optimize generated relations (\eg, attachment and support) from Sec.\ref{ssec:init_estim_relation}. 
As the name suggests, the magnetic loss encourages two opposite surfaces of the attached objects to be close to each other without violation.
Practically, we shoot a set of probing rays from one object surface plane to another with the shooting direction from the generated relations (Sec.~\ref{ssec:init_estim_relation}) and compute ray-to-surface distances via sphere tracing.
Then, as illustrated in Fig.~\ref{fig:magnetic_loss}, we define two distances to diagnose surface-surface relations: 1) attachment distance $d_{a}$ which measures the surface distance between two objects by summing up the distances of partial nearest intersections while ignoring far-away intersections, 2) violation distance $d_{v}$ which indicates potential violation of two objects by summing up all the violation part of the surfaces.
To this end, we defined magnetic loss as:
\begin{equation}
    L_{mag}= \frac{1}{K}\sum_{k=1}^K \max(d_{a}^k,0) + \max(d_{v}^k,0).
\end{equation}
Please refer to the supplementary materials for more details.

\PAR{Physical Violation Loss.}
To mitigate physical occlusions that the magnetic loss does not cover (\eg, chair under the desk), inspired by Zhang \etal~\cite{im3d}, we add a physical violation loss based on the neural SDF model.
Different from Zhang \etal~\cite{im3d} which uniformly samples points inside the bounding boxes, we only sample points on the visible surface with outside-in sphere tracing, so as to make the optimization more efficient when two objects only collide partially at a shallow level.
The physical violation loss is defined by punishing $P$ surface points for each object $k$ when an object's points lie inside its $O$ neighbor objects by querying the corresponding SDF surface model $F_{\text{SDF}}$ as:
\begin{equation}
    L_{vio}=\sum_{k=1}^K \sum_{o=0}^{O}\sum_{p=1}^{P}\min(\text{SDF}({\mathbf{x}_k}_p)_o+\epsilon,0),
\end{equation}
where $o=0$ denotes the scene background, and we set $\epsilon=0.025$, $O=3$ and $P=1000$ in our experiment.

\noindent\PAR{Gravity Direction Loss.}
In real-world scenarios, many furniture like beds and tables only rotate around the gravity direction (\ie, rotation uncertainty only on the yaw angle). So we also add the gravity-direction energy term to the physical constraints for those objects as:
\begin{equation}
    L_{g} = \sum_{j=1}^{K} \similarity(\hat{\mathbf{R}}^k \mathbf{g}, \mathbf{g}),
\end{equation}
where $\mathbf{g}=[0, 0, 1]^{\top}$ is the gravity direction .

The overall physical loss is defined as: $L_{phy} = L_{mag} + L_{vio} +  L_{g}$.

\subsubsection{Holistic Optimization}
\hfill\\
In holistic optimization, we seek for per-object poses $\hat{\mathbf{T}}^k$, object and background appearance codes $\bm{l}_{a}^k$ and light codes $\bm{l}_{l}^k$ that satisfy the input panoramic image.
To fulfill this goal, we jointly optimize photometric loss, observation loss and physical constraint losses at the online stage, as:
\begin{equation}
    L = \lambda_{pho} L_{pho} + \lambda_{obs} L_{obs} + \lambda_{phy} L_{phy}.
\end{equation}
We use $\lambda_{pho}=1$, $\lambda_{lbs}=100$, and $\lambda_{phy}=1$ in our experiment.
The total optimization takes about 10-15 minutes (depending on the frequency of object occlusions) for a panoramic image with 500 iterations on an Nvidia RTX3090-24G graphics card.
More discussion of the time-consuming and the possible improvement at the online stage can be found in Sec.~\ref{sec:conclusion}.

\section{Experiments}
\noindent In this section, we first compare our scene arrangement prediction with DeepPanoContext~\cite{deeppanocontext} and evaluate the scene lighting adaptation ability both quantitatively and qualitatively.
Then, we perform ablation studies to analyze the design of our framework.
Finally, we demonstrate the applicability of our method on scene touring, scene illumination interpolation, and scene editing.

\subsection{Dataset}
\PAR{iG-Synthetic.}
We use iGibson~\cite{igibson} simulator to synthesize labeled images with depth, segmentation and 3D bounding boxes for training and testing.
For training object-centric models for identification, pose estimation or neural rendering, we generate 360\textdegree~views around each object (similar to Realistic Synthetic 360\textdegree~in NeRF~\cite{nerf}).
For the background scenes, we leverage the toolbox from Zhang \etal~\cite{deeppanocontext} to generate panoramic views of the iGibson scenes.
Since many rooms in iGibson are either too empty (\eg, bathroom and storage-room) or filled with fixed stuff (\eg, basin and oven in kitchen), we thus select four representative scenes (\ie, bedroom, lobby, child's room and home office) which already covers most of the movable object types in the dataset.

\PAR{Fresh-Room.}
To demonstrate the efficacy in real-world scenes, we create a new dataset named Fresh-Room, which contains RGB-D posed training images for 5 objects and the room background captured by iPad Pro.
We also capture multiple panoramic testing images under 4 different setups with varying arrangements and lighting conditions using a 360\textdegree camera (Insta360 ONE-R).
We utilize the SfM system with mesh reconstruction~\cite{colmap,kazhdan2006poisson} and ARKit metadata\footnote{https://developer.apple.com/documentation/arkit/arcamera} to recover camera poses with real-world scale and obtain 2D segmentation by projecting annotated labels from 3D meshes for training data.

\subsection{Scene Arrangement Prediction}
\begin{figure}[t!]
    \centering
    \includegraphics[width=1.0\linewidth, trim={0 0 0 0}, clip]{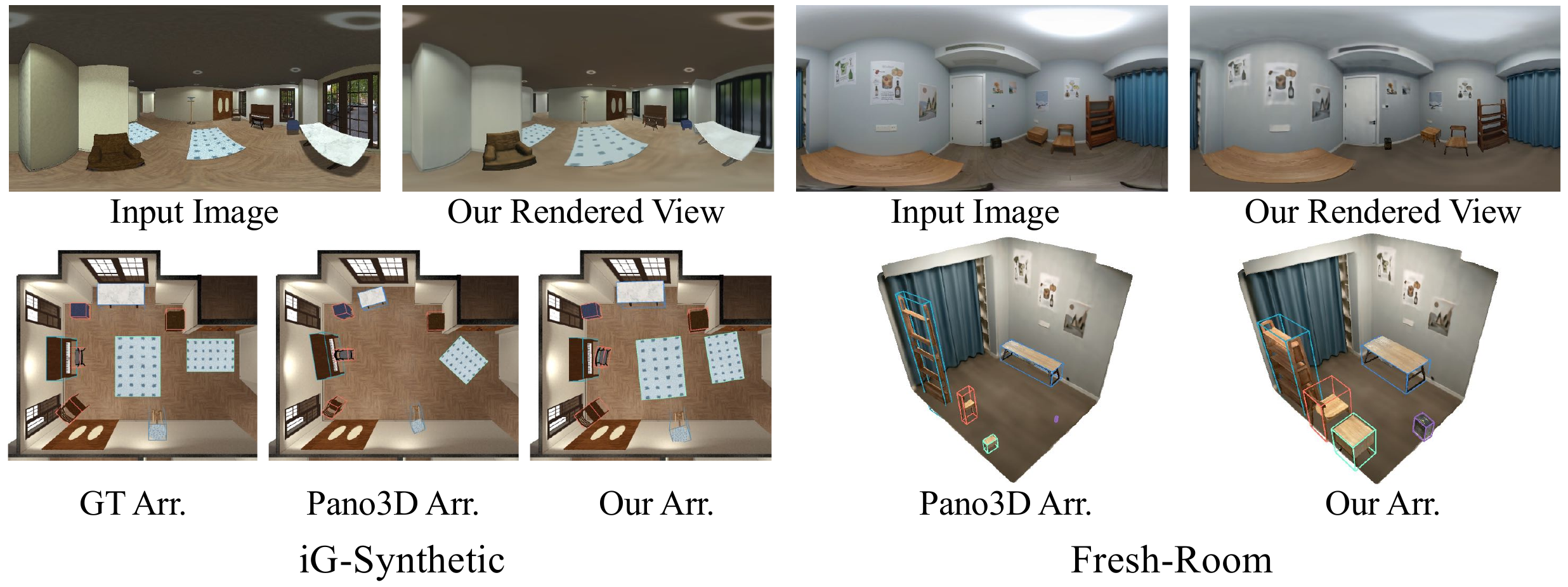}
    \caption{
    Scene arrangement visualization with textured meshes, where objects are scaled by the estimated bounding boxes.
    Note that for the Fresh-Room, we use the object meshes extracted from our neural implicit renderer.
    }
    \label{fig:scene_arrangement}
\end{figure}

\noindent We first evaluate scene arrangement prediction on iG-Synthetic dataset and our Fresh-Room dataset.
For iG-Synthetic dataset, we reorganize the scene arrangement following room examples~\cite{igibson}, producing unseen arrangements for four scenes, and synthesizing testing data with $5$ unseen indoor illuminations based on the iGibson PBR engine.
Since there is no amodal scene understanding approach for comparison, we take DeepPanoContext~\cite{deeppanocontext} (Pano3D) as a reference, which is a SOTA method for general holistic 3D scene understanding with a panoramic input. The Intersection over Union (IoU), Average Rotation Error (ARE) and Average Position Error (APE) are used as evaluation metrics.
As demonstrated in Fig.~\ref{fig:scene_arrangement} and Tab.~\ref{tab:scene_arrangement}, our method consistently achieves better scene arrangement prediction quality both quantitatively and qualitatively 
under a closed scene, where the furniture like shelf and piano are faithfully placed with accurate size, while the general scene understanding method (DeepPanoContext) struggles to produce satisfying results (\eg, the desk and the piano are tilted in iG-Synthetic, and the size of shelf and nightstand are distorted in Fresh-Room).
This experiment shows that our amodal 3D understanding approach makes a further step towards a perfect 3D understanding, which benefits from the offline preparation stage.

\begin{table}[tb]
\caption{Quantitative evaluation on scene arrangement prediction.}
\resizebox{1\linewidth}{!}{
\begin{tabular}{lcccccc}
\toprule
\multicolumn{1}{c}{\multirow{2}{*}{Scene}} & \multicolumn{3}{c}{DeepPanoContext} & \multicolumn{3}{c}{Ours} \\ \cmidrule(lr){2-4} \cmidrule(lr){5-7} 
\multicolumn{1}{c}{} & IoU (\%) $\uparrow$ & ARE (\textdegree) $\downarrow$ & APE (cm) $\downarrow$ & IoU (\%) $\uparrow$ & ARE (\textdegree) $\downarrow$ & APE (cm) $\downarrow$ \\ \midrule
Lobby & 26.34 & 52.80 & 23.47 & \textbf{44.48} & \textbf{33.99} & \textbf{10.08} \\
Bedroom & 40.61 & 30.33 & 11.43 & \textbf{55.42} & \textbf{25.88} & \textbf{9.14} \\
Child's Room & 27.76 & 34.64 & 17.78 & \textbf{48.63} & \textbf{29.57} & \textbf{9.08} \\
Home Office & 38.04 & 27.47 & 11.45 & \textbf{47.91} & \textbf{19.95} & \textbf{6.46} \\
\midrule
Average & 33.19 & 36.31 & 16.03 & \textbf{49.11} & \textbf{27.35} & \textbf{8.69}\\ 
\bottomrule
\end{tabular}
}
\label{tab:scene_arrangement}
\end{table}

\subsection{Scene Lighting Adaptation}
\label{ssec:light_ada}

\begin{figure}[t!]
    \centering
    \includegraphics[width=1.0\linewidth, trim={0 0 0 0}, clip]{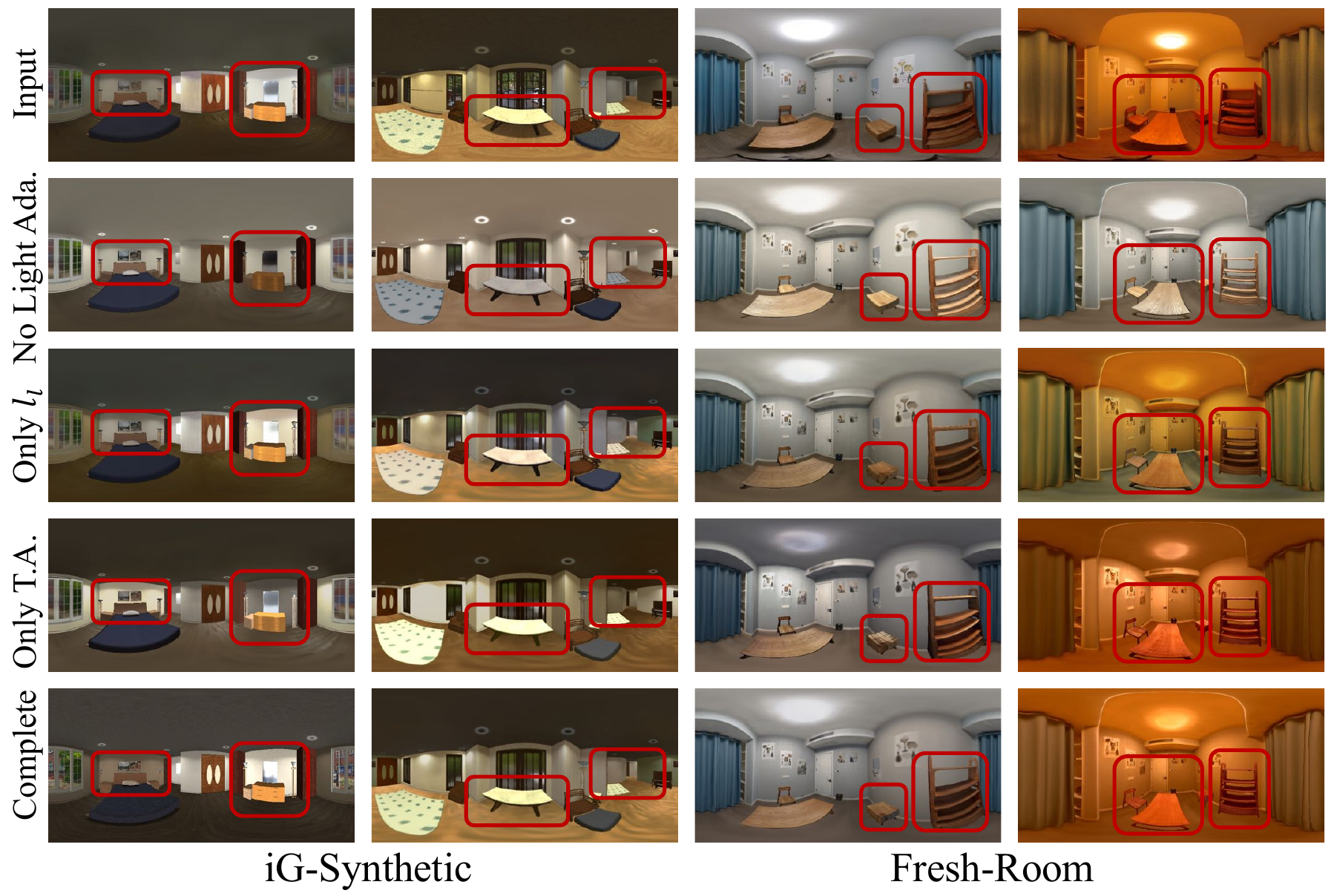}
    \caption{
    \textbf{Lighting Adaptation.}
    We can adapt lighting condition to the input panorama with light code optimization ($\bm{l}_l$) and tone adjuster (T.A.).
    }
    \label{fig:light_adapt}
\end{figure}

\noindent 
Since we decouple lighting variation implicitly in a latent space ($\bm{l}_{l}$) and explicitly via tone adjuster, hence we can adjust the renderer to fit the lighting condition at test time.
As shown in Fig.~\ref{fig:light_adapt}, the input images (first row) have dramatically different lighting variations such as local highlight and global warm light.
When the lighting adaptation is disabled (second row), the rendered results are close to the pre-captured training views, where the rendered furniture comes up with inconsistent shininess (\eg, the shelf in the third column are inevitably brighter than the real image).
By introducing implicit light code optimization (third row), the rendered scenes are closer to the input ground-truth but struggles to adapt to the extremely warm light in the last column, where the global tone has turned yellow but the floor color and the curtain color are distorted.
By enabling the tone adjuster (fourth row) only, we can also handle a certain degree of lighting variation (\eg, carpet and desk lit by yellow light in the second row, warm light in fourth column), but fails to adapt the local lighting  variation (\eg, scene background partially lit by strong light in first column).
When the tone adjuster and the light code optimization are both enabled, we successfully render images with local highlight and global consistent tone, and also achieve the best metric performance as demonstrated in Tab.~\ref{tab:light_adapt}.
We believe that the tone adjuster effectively reduces the burden of latent space optimization, and the combination of explicit and implicit optimization enhances the lighting adaptation ability.

\begin{table}[tb]
\caption{Ablation study of light code optimization ($l_l$) and tone adjuster (T.A.) of light adaptation.}
\resizebox{1\linewidth}{!}{
\begin{tabular}{lcccccc}
\toprule
\multicolumn{1}{c}{\multirow{2}{*}{Config.}} & \multicolumn{3}{c}{iGibson-Synthetic} & \multicolumn{3}{c}{Real-Room} \\ \cmidrule(lr){2-4} \cmidrule(lr){5-7} 
\multicolumn{1}{c}{} & PSNR $\uparrow$ & SSIM $\uparrow$ & LPIPS $\downarrow$ & PSNR $\uparrow$ & SSIM $\uparrow$ & LPIPS $\downarrow$ \\ \midrule
No light ada.  & 13.939  & 0.674   &  0.359  &  13.832  & 0.592   &  0.480  \\ 
Only T.A.  & 16.836   & 0.732   & 0.334  &  17.302  & 0.615  &  0.318  \\
Only $l_l$ &  23.015  &  0.809  &  0.302  & 19.471   & 0.669   & 0.345   \\
Complete & \textbf{24.073} & \textbf{0.821} & \textbf{0.279} & \textbf{21.341} & \textbf{0.683} & \textbf{0.288} \\
\bottomrule
\end{tabular}
}
\label{tab:light_adapt}
\end{table}

\subsection{Ablation Studies}

\PAR{Data Augmentation.}
We first analyze the effectiveness of our data augmentation with compositional neural rendering for object prediction.
Specifically, we use the ODN network~\cite{deeppanocontext,total3d} as a baseline and fine-tune on the labeled panoramic images rendered by our neural implicit renderer.
The results in Tab.~\ref{tab:ablation} show that our proposed data augmentation improves the object prediction quality on the rotation error and IoU (first two rows),
and also boosts the performance of scene arrangement prediction in holistic optimization (third row and last row).

\PAR{Various Constraints.}
We then inspect the efficacy of various constraints in our holistic optimization, including photometric constraint in Sec.~\ref{ssec:photo_optim}, observation constraint in Sec.~\ref{ssec:obs_optim} and physical constraint in Sec.~\ref{ssec:physical_optim}.
Note that we exclude lighting adaptation as this process is mainly for better rendering quality and performed after the object pose optimization (see supplementary for more details).
As shown in Tab.~\ref{tab:ablation} (last five rows), all these loss terms improve the overall scene arrangement quality.
Furthermore, we evaluate the performance when only a bare photometric loss is enabled.
It is clear that without physical 
and observation constraints, a simple photometric loss is prone to be unstable in such cluttered scenes.

\PAR{Safe-Region Volume Rendering.}
We also compare our proposed safe-region volume rendering with the classical volume rendering pipeline (\eg, used by iNeRF~\cite{inerf}) in the pose optimization task with an Nvidia RTX-3090-24GB graphics card.
During the experiment, we optimize object poses by jointly rendering background scenes and target objects with back-propagation,
and report the GPU memory usage by varying the number of sample rays and target objects.
As shown in Tab.~\ref{tab:safe_region}, our proposed strategy significantly reduces the number of network queries and GPU memory consumption and can simultaneously optimize 10 objects, while the classical volume rendering fails due to out of memory.
We verify the impact on the pose estimation quality in the supplementary material, which shows that this strategy maintains similar pose convergence performance as classical volume rendering.

\begin{table}[tb]
\caption{Ablation studies for the data augmentation and various constraints.
}
\resizebox{1.0\linewidth}{!}{
\begin{tabular}{lcccccc}
\toprule
\multicolumn{1}{c}{\multirow{2}{*}{Config.}} & \multicolumn{3}{c}{iGibson / Lobby} & \multicolumn{3}{c}{iGibson / Bedroom} \\ \cmidrule(lr){2-4} \cmidrule(lr){5-7} 
\multicolumn{1}{c}{} & IoU (\%) $\uparrow$ & ARE (\textdegree) $\downarrow$ & APE (cm) $\downarrow$ & IoU (\%) $\uparrow$ & ARE (\textdegree) $\downarrow$ & APE (cm) $\downarrow$ \\ \midrule
ODN &  10.75  & 58.46  & 38.29  & 16.30  & 33.89  & 26.33 \\
ODN w aug.  & 13.01  & 39.17  & 45.50  & 20.43  & 32.12  & 23.70   \\ 
\midrule
w/o aug. & 37.01 & 50.93 & 13.53 & 56.99 & 38.20 & 8.27  \\ 
w/o $L_{pho}$  & 42.34 & 33.92 & 10.17 & 51.54 & 26.10 & 10.31  \\ 
w/o $L_{phy}$  & 16.93 & 43.81 & 35.61 & 23.65 & 37.69 & 23.31   \\ 
w/o $L_{obs}$  & 33.45 & 34.88 & 16.76 & 44.91 & 24.48 & 15.40   \\ 
only $L_{pho}$  & 21.05 & 42.24 & 34.00 & 20.57 & 34.22 & 28.99  \\
\midrule
Complete & 44.48 & 33.99 & 10.08 & 55.42 & 25.88 & 9.14 \\
\bottomrule
\end{tabular}
}
\label{tab:ablation}
\end{table}

\begin{figure}[t!]
    \centering
    \includegraphics[width=1.0\linewidth, trim={0 0 0 0}, clip]{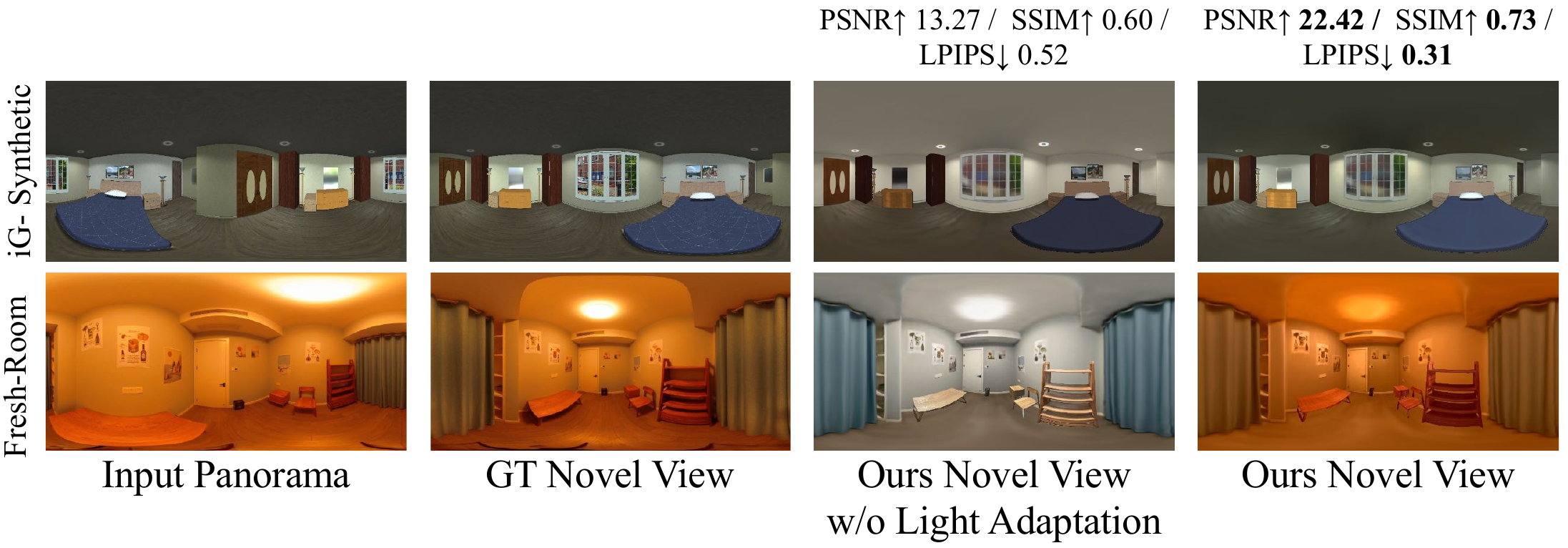}
    \caption{
    \textbf{Free-viewpoint scene touring} on the iG-Synthetic dataset and Fresh-Room dataset.
    }
    \label{fig:scene_rerendering}
\end{figure}

\begin{table}[tb]
\caption{Ablation study for the safe-region volume rendering (S.R.) with different number of target objects and rays. 
Bg.+1/10 Obj. denotes joint rendering with background and 1 or 10 objects.
$\times$ denotes out of memory.}
\resizebox{1\linewidth}{!}{
\begin{tabular}{rcccccccc}
\toprule
Config. & S.R./Bg.+1 Obj. & no S.R./Bg.+1 Obj. & S.R./Bg.+10 Obj. & no S.R./Bg.+10 Obj. \\
\midrule
\# Rays & \multicolumn{4}{c}{\# Query / GPU Memory} \\
\midrule
256 & \textbf{5.2M / 2.5G} & 19.7M / 6.4G & \textbf{18.1M / 3.3G} & 51.4M / 7.8G \\
512 & \textbf{23.0M / 3.4G} & 86.4M / 11.1G & \textbf{70.8M / 4.3G} & 220.8M / 13.6G \\
1024 & \textbf{91.9M / 5.2G} & 346M / 20.2G & \textbf{287.0M / 6.4G} & $\times$ \\
2048 & \textbf{274.0M / 6.9G} & $\times$ & \textbf{1170.2M / 10.5G} & $\times$ \\
\bottomrule
\end{tabular}
}
\label{tab:safe_region}
\end{table}

\subsection{Free-viewpoint Scene Touring}
\noindent Once we resolve the scene arrangement and scene lighting condition, it is feasible to re-render the room in any arbitrary view, which enables virtual touring of the room.
To inspect the rendering quality for this task, we conduct a scene re-rendering experiment by fitting input images (first column in Fig.~\ref{fig:scene_rerendering}) and render another view with the fitting results.
Thanks to our neural scene representation, the rendered novel views (last column in Fig.~\ref{fig:scene_rerendering}) vividly reproduce scene appearance and lighting conditions (\eg, local highlight and global warm light) 
of the corresponding ground-truth novel views (second column in Fig.~\ref{fig:scene_rerendering}).
As a comparison, when ablating the lighting adaptation from the representation, we can still achieve realistic novel view rendering results, but the specific lighting conditions (\eg, local highlight and warm tone) are no longer kept (third column in Fig.~\ref{fig:scene_rerendering}), which also results in lower metric performances.

\begin{figure}[t!]
    \centering
    \includegraphics[width=1.0\linewidth, trim={0 0 0 0}, clip]{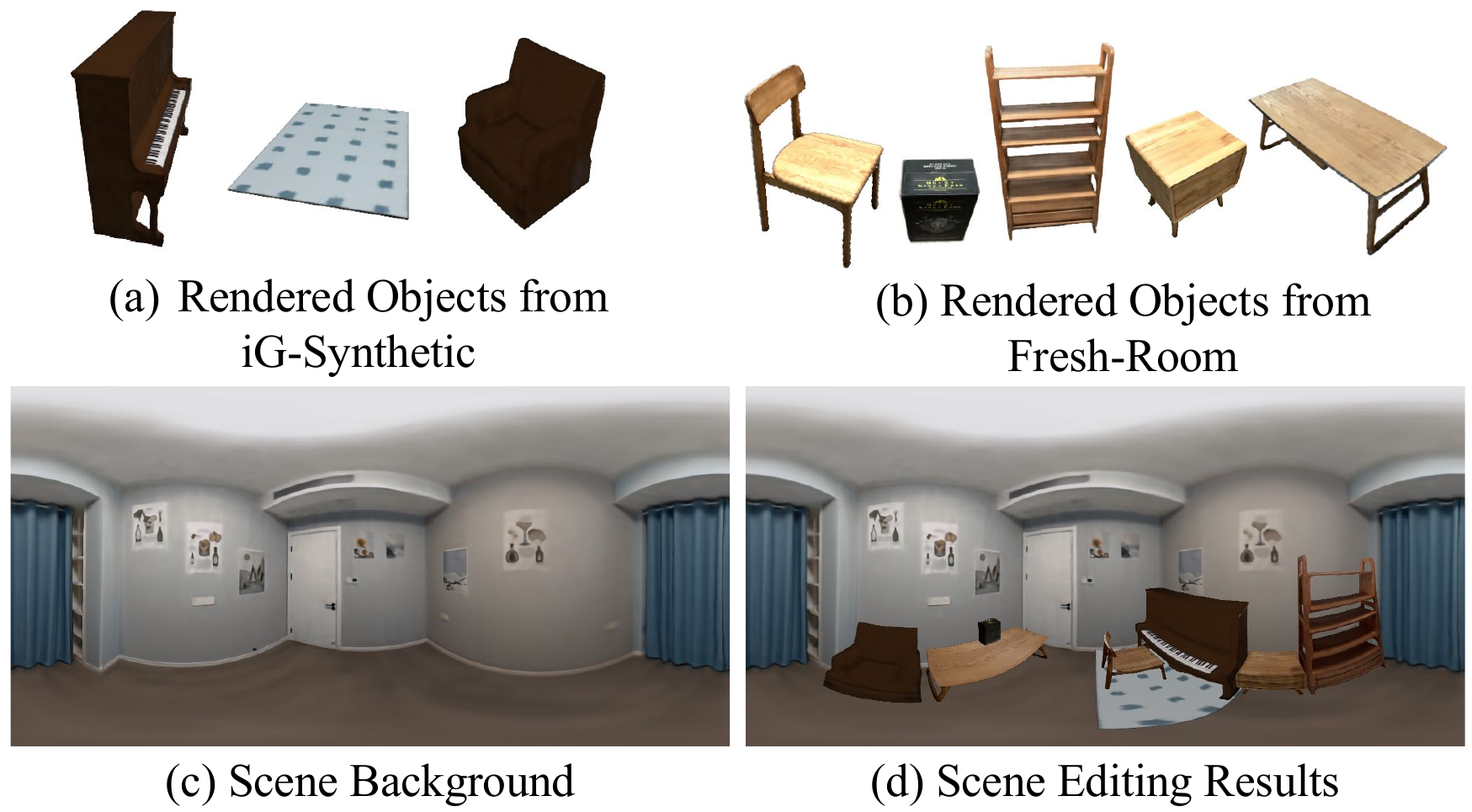}
    \caption{
    \textbf{Scene Editing.} We insert virtual objects (piano, sofa chair and carpet) into the real-world.}
    \label{fig:scene_edit}
\end{figure}

\begin{figure}[t!]
    \centering
    \includegraphics[width=1.0\linewidth, trim={0 0 0 0}, clip]{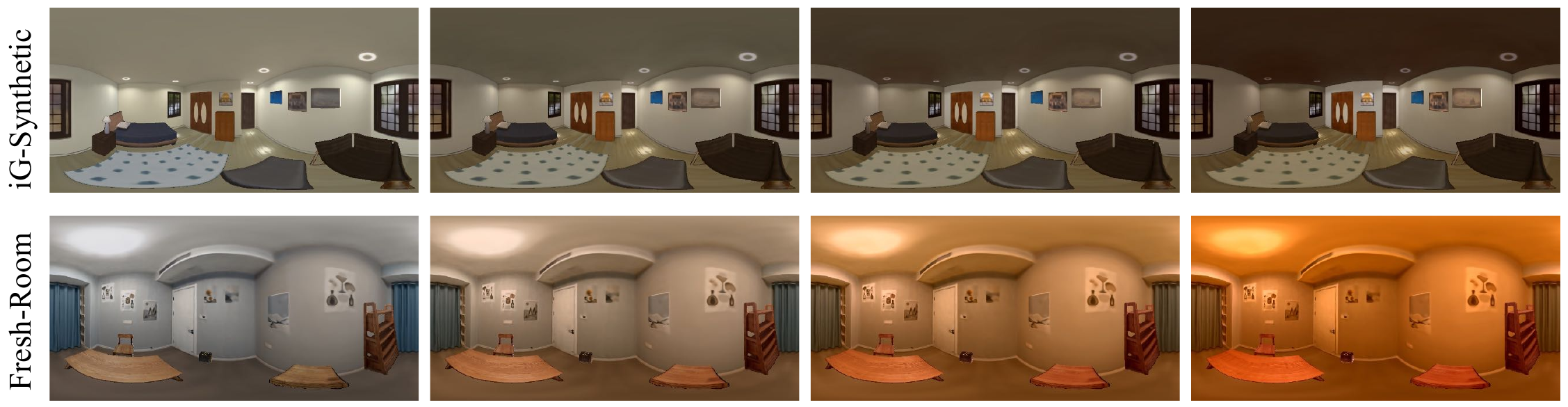}
    \caption{
    \textbf{Illumination interpolation} on the iG-Synthetic dataset and Fresh-Room dataset.
    }
    \label{fig:light_interp}
\end{figure}

\subsection{Scene Editing \& Illumination Interpolation}

\noindent Since our neural implicit renderer has already learned to render the scene background and objects, we can easily edit or composite novel scenes upon this.
As shown in Fig.~\ref{fig:scene_edit}, we perform scene editing by inserting virtual objects learned from iG-Synthetic into the real scene Fresh-Room, and the rendered image of novel view demonstrates correct space relationship with seamless object-object occlusion.
We also conduct the illumination interpolation experiment in Fig.~\ref{fig:light_interp}, where the scene lighting temperature can be naturally turned from day to night (first row), or from cold to warm (second row).

\section{Conclusion}
\label{sec:conclusion}
\noindent We propose a novel 3D scene understanding and rendering paradigm for closed environments.
Given a single panoramic image as input, our method can reliably estimate 3D scene arrangement and the lighting condition via a holistic neural-rendering-based optimization framework.
The proposed method also enables free-viewpoint scene touring and editing by changing illumination or objects' placement.
Despite the novel capabilities provided by our method, it still has its limitations. 
First, since we assume the neural implicit models are pre-built, our method cannot handle the cases with unobserved objects.
Second, the computational efficiency of the online stage is currently not ready for real-time performance, which is due to the intensive network queries of MLPs.
There are some existing approaches accelerating neural volumetric rendering from ~0.06FPS to ~200FPS, e.g., by using local-bounded representation~\cite{reiser2021kilonerf}, cached coefficients~\cite{garbin2021fastnerf,yu2021plenoxels}, or multiresolution voxel-hashing~\cite{muller2022instant}, and they can be applied for real-time rendering and fast optimization, which is a promising future direction.
Third, the proposed method still cannot handle deformed/recolored objects and extremely harsh lighting that severely violates photometric consistency, or render transparent surfaces and fine-grained light effects like shadows and indirect illumination.
Finally, our lighting augmentation is not well-defined for glossy materials like mirrors and glasses, which can be improved by introducing material estimation in the future.

\section{Acknowledgments}
We thank Hanqing Jiang and Liyang Zhou for their kind help in object reconstruction.
This work was partially supported by NSF of China (No.~62102356, No.~61932003) and Zhejiang Lab (2021PE0AC01).

\bibliographystyle{ACM-Reference-Format}
\bibliography{main}

\clearpage

\appendix

\renewcommand\thesection{\Alph{section}}
\renewcommand\thetable{\Alph{table}}
\renewcommand\thefigure{\Alph{figure}}

{\noindent \Large \bf Supplementary Material\par}

\vspace{0.5em}

\noindent In this supplementary material,
we will describe more details of our framework including object prediction in Sec.~\ref{sec:object_pred}, neural implicit renderer in Sec.~\ref{sec:neural_implicit_renderer}, and holistic optimization in Sec.~\ref{sec:holistic}.
Besides, we also provide more discussion in Sec.~\ref{sec:discussion} and more experimental results in Sec.~\ref{sec:expr}.

\section{Details of Object Prediction}
\label{sec:object_pred}

\subsection{Object Detection}

\PAR{Implementation Details.}
We adopted the Object Detection
Network (ODN) from~\cite{deeppanocontext, total3d} to predict the initial object poses.
The vanilla ODN extracts the appearance feature with ResNet-18 from 2D detection, and encodes the relative position and size between 2D object boxes into geometry features.
Differently, to generalize better with our data augmentation where the objects are randomly placed in the neural rendered views, we replace the object relative features with object sizes and positions of themselves.
Even though we do not rely on object relation knowledge, we still obtain performance gains in the object pose estimation thanks to the data augmentation.

\PAR{Training and Testing Details.}
The ODN is fine-tuned on the labeled panoramic images synthesized by our neural implicit renderer.
We train the network for 10 epochs with a batch size of 6 in one hour on a single Nvidia RTX-3090-24G graphics card.
During the testing stage of the Fresh-Room dataset, we also pre-process the input image with automatic white-balance correction~\cite{wb_srgb} to reduce the impact of lighting variation on 2D detection, which is also applied to DeepPanoContext~\cite{deeppanocontext} for fair comparison.
Moreover, according to the statistics, the testing data of iGibson-Synthetic contains frequent occlusions (with about 16\% in Lobby, 35\% in Bedroom, 31\% in Child's room, and 40\% in Office room).

\subsection{Object Identification (ReID)}
During the ODN training phase, we generate extensive labeled panoramas by compositional neural rendering with the implicit neural renderer.
Each 2D object boxes in the equirectangular images is represented as a Bounding FoV (BFoV)~\cite{obj_detect_panorama}, which is defined with the latitude and longitude of the center and the horizontal and vertical field of view.
We crop out the objects in each BFoV from the panoramas, warp them into perspectives
and then save the NetVLAD features of these perspectives with their corresponding object ids as our retrieval database.

In the identification phase, we compute the NetVLAD features of each 2D detection boxes, and then find the one with the highest similarity in the database. We remove the query boxes with a match similarity threshold less than $s=0.25$.
The remaining query boxes are identified as the most similar objects in the database.

\section{Details of Neural Implicit Renderer}
\label{sec:neural_implicit_renderer}

\subsection{Model Architecture}
We adopt NeuS~\cite{neus} as the network backbone for the neural implicit renderer.
As introduced in Sec.~\textcolor{black}{3.1.1}, we append 64-dimension object codes $\bm{l}_{obj}$ to the input of the SDF surface model $F_{\text{SDF}}$, so as to control the visibility of a certain object identity.
For the radiance model $F_{\text{R}}$, we add 16-dimension appearance code $\bm{l}_{a}$ to handle the sensing variation on the Fresh-Room dataset, and add 16-dimension light code $\bm{l}_{l}$ to learn the latent space of illumination variation.
Besides, to enhance the rendering texture details on the Fresh-Room dataset, we also extend the positional encoding~\cite{nerf} of the spatial coordinate to 10 in $F_{\text{R}}$.

\subsection{Training Details}
\begin{figure}[t!]
    \centering
    \includegraphics[width=1.0\linewidth, trim={0 0 0 0}, clip]{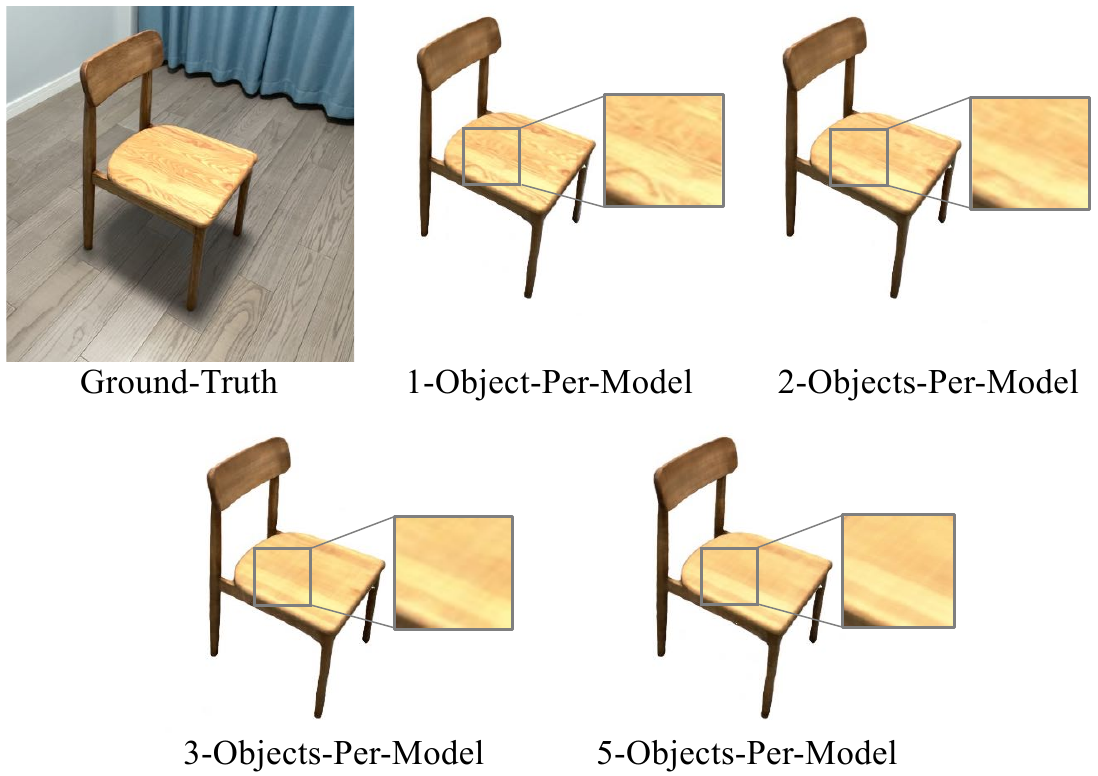}
    \caption{
    We visualize the rendered chairs of the Fresh-Room dataset with varying model training settings (from 1-model-per-object to 5-model-per-object).
    We observe that a single neural model can encode 5 real-world objects, but may lose some texture details.
    }
    \vspace{-1.0em}
    \label{fig:supp_multi_obj_per_model}
\end{figure}

Following Yang \etal~\cite{object_nerf}, we use color, depth, and object losses to supervise the neural implicit renderer.
To adapt the geometric initialization in the SDF surface model to the inside-out reconstruction for room background, we reverse the sign of the bias in the last MLP~\cite{igr} during network initialization, so the initial surface normal is facing towards the scene center.
Due to the limitation of network capacity, we do not learn every objects in one MLP, but encode 4 or 5 objects per model for the iG-Synthetic dataset and 1 object per model for the Fresh-Room dataset.
Note that we use a different model setting for real/synthetic data, which is a trade-off between model capacity, training time and rendering quality. 
Since the appearance detail of Fresh-Room is richer than iGibson, we thus deliver the best visual quality with a per-model-per-object setting.
Please refer to the Fig.~\ref{fig:supp_multi_obj_per_model} for the visualization of rendering quality of the Fresh-Room dataset under different model settings.
During training, we adopt the Adam optimizer with an initial learning rate of 0.0005 and a polynomial decay schedule with the power of 2.
The training process for each neural model takes about 10 hours with a batch size of 1024 rays on a single NVidia RTX-3090-24G graphics card.

\section{Details of Holistic Optimization}
\label{sec:holistic}

\subsection{Details of Relation Generation}
As introduced in Sec.~\textcolor{black}{3.2.1}, we use the object meta information from object prediction and geometric cues (extracted bounding boxes, ray intersection distance and normal) to infer a set of relations for physical constraints optimization, including object-object support, object-wall attachment and object-floor support.
In practice, we put the object with the initial pose prediction and detect surface-to-surface distances with a set of probing rays (similar to that in magnetic loss) along with several pre-defined directions (downward for support relations, and four perpendicular directions to the bounding box surfaces for attachment relation) and find the relation candidates with distances smaller than 0.5m.
For the attachment relation candidates, we filter the case that the surface is not opposite to the walls by ensuring intersection point normal almost opposite to the probing ray shooting directions, \eg, we only allow the back face of a cabinet to attach to the wall rather than the front face.
Finally, we apply custom rules to further filter out unexpected relations and ensure several definite relations (\eg, bed, shelf, nightstand and wall-pictures are always attach to wall; bed, chairs and bottom cabinet are always attach to the floor; PC monitor, wall-pictures and mirrors should not attach to the floor).
During the holistic optimization, we periodically regenerate relations per 50 iterations, so the potential relations that cannot be found at the beginning can be supplemented when object poses are better converged.

\subsection{Details of Photometric Constraint Optimization}
We empirically sample 1024 rays with the resolution of 320$\times$160 when optimizing photometric loss.
Because the equirectangular images are squeezed on vertical center and stretched on top/bottom poles, we draw a biased sampling with more rays near the vertical center and less when close to the poles.
For the ray samples on the object masks, we adopt the region-based sampling~\cite{inerf} by dilating object masks with 5$\times5$ kernels.
In practice, we first optimize object poses and then optimize appearance and light codes (with zero-initialized) while freezing object poses, since we found the convergence of poses and lighting is more stable in this way.

\subsection{Details of Safe-Region Volume Rendering}
We use a safe-region volume rendering strategy in photometric constraint optimization, where we only sample a few points near the surface in volume rendering.
Practically, for each ray, we first perform an efficient sphere tracing to find the ray-intersection point, where the ray-to-surface distance is denoted as $d_{\text{surf}}$.
Then, we empirically bound NeRF's hierarchical sampling near-far range to $d_{\text{surf}}-0.05$ and $d_{\text{surf}}+0.1$, and only sample 8 points for ``coarse'' sampling and 16 points for ``fine''  sampling.
Besides, we also analysis the impact to the pose convergence w.r.t. the number of sample points of this strategy in Sec.~\ref{ssec:analysis_safe_region}.

\subsection{Details of Magnetic Loss}
We propose magnetic loss in Sec.~\textcolor{black}{3.2.1}, which faithfully attach two opposite surfaces for two objects that have been marked as attachment or support relations.
As explained in the main paper, we use several probing rays to determine attachment distance $d_a$ and violation distance $d_v$.
In our experiment, to handle the potential violations that hinders sphere tracing (\eg, objects inside the wall), the starting points of the probing rays have always been back-offset for a certain distance $d_{\text{back}}$. The $d_{\text{back}}$ is starting from 0.5m, and is gradually increased until the sphere-tracing finds the intersection or $d_{\text{back}}$ exceeds the preset upper-bound (1.5m).
For the surface plane that the ray array comes from, we use a similar ray-intersection test to find the surface points for attachment relations (\eg, looking for the side of the bed and ignoring the hollow part). 
Besides, for support relations, we directly use extracted bounding box bottom plane (\eg, using bed's bounding box bottom as surface plane, since the bed legs or cabinet legs might be too tiny for the probing-ray-based intersection detection).

\section{More Discussions}
\label{sec:discussion}

\subsection{Applicability for Real-World Scenes}
When running with complex real-world scenes, since we use SDF-based constraints instead of bounding boxes, the holistic optimization also supports cluttered objects (with gravity loss omitted for certain object types).
However, we need to refactor the object detector~\cite{HRNet,tan2020efficientdet} and relation generator~\cite{xu2019densephysnet,xu2022umpnet} to support cluttered objects like books and pens.
Besides, for the rooms containing windows with dynamic landscapes, we recommend applying masks to the images.
And it might also be possible to simplify the offline stage (\eg, using domestic robots instead of hand-held capturing) when deploying to real applications.
Note that our method supports both panoramic and non-panoramic images by simply switching the ray generation from spherical to perspective fashion. Empirically, fewer objects will be visible if using non-panoramic images, and we may need multiple views to understand the whole scene.

\subsection{Scale with Number of Objects}
The computation resources (e.g., GPU memory) of our method grow linearly with the number of different objects, and we've tested 13 objects in the experiments.
Besides, for the scene with massive similar objects (\eg, a classroom scene with 20-50 similar desks), we can reuse rendering models to save memory.

\subsection{Offline-Stage Efforts}
Generally, we take ~30 minutes to capture the Fresh-Room data, and spend ~15 minutes to annotate all the object meshes. The other process (e.g., reconstruction and training) can be automated and trained in parallel within ~20 hours. Note this needs to be run only once.

\subsection{Objects with the Same Shape but Different Colors}
Sometimes, we may have two objects with identical shape but just different colors (\eg, a red chair and a blue chair).
Note that our method does not handle such cases, as this is not available in our dataset. In the future, we can disentangle color and geometry by simply enforcing shape coherence (\eg, Edit-NeRF~\cite{liu2021editing}) to benefit applications when handling mass-produced furniture like IKEA.

\section{More Experimental Results}
\label{sec:expr}

\subsection{Performance of Re-Identification}

\begin{table}[tb]
\caption{ReID Performance on the iG-Synthetic dataset.
}
\resizebox{1\linewidth}{!}{
\begin{tabular}{lccccc}
\toprule
Scene & Lobby & Bedroom & Child's Room & Home Office & Average \\ 
\midrule
Precision & 91.84 & 78.79 & 84.48 & 88.72 & 85.96 \\ 
\midrule
Recall & 86.33 & 76.42 & 75.00 & 79.09 & 79.21 \\ 
\bottomrule
\end{tabular}
}
\newline
\label{tab:reid}
\end{table}

We evaluate the performance of our ReID module on the iG-Synthetic dataset with four scenes (bedroom, lobby, child’s room and home office).
The scenes are re-organized to unseen arrangements and unseen indoor illuminations, as described in the original paper.
We adopt two commonly-used metrics, \ie, precision and recall, to measure the ReID performance, as shown in Table~\ref{tab:reid}.

Although the panoramic image can capture all the unoccluded objects, there are still some objects that are too far away or severely squeezed due to equirectangular projection, which are hard to be detected and affects the performance of ReID.
These missing detection can be handled by introducing positional prior in our future work, where we can assume the furniture arrangement is only gradually changing for daily lives.

\subsection{Analysis of Safe-Region Volume Rendering}
\label{ssec:analysis_safe_region}

\begin{figure}[t!]
    \centering
    \includegraphics[width=1.0\linewidth, trim={0 0 0 0}, clip]{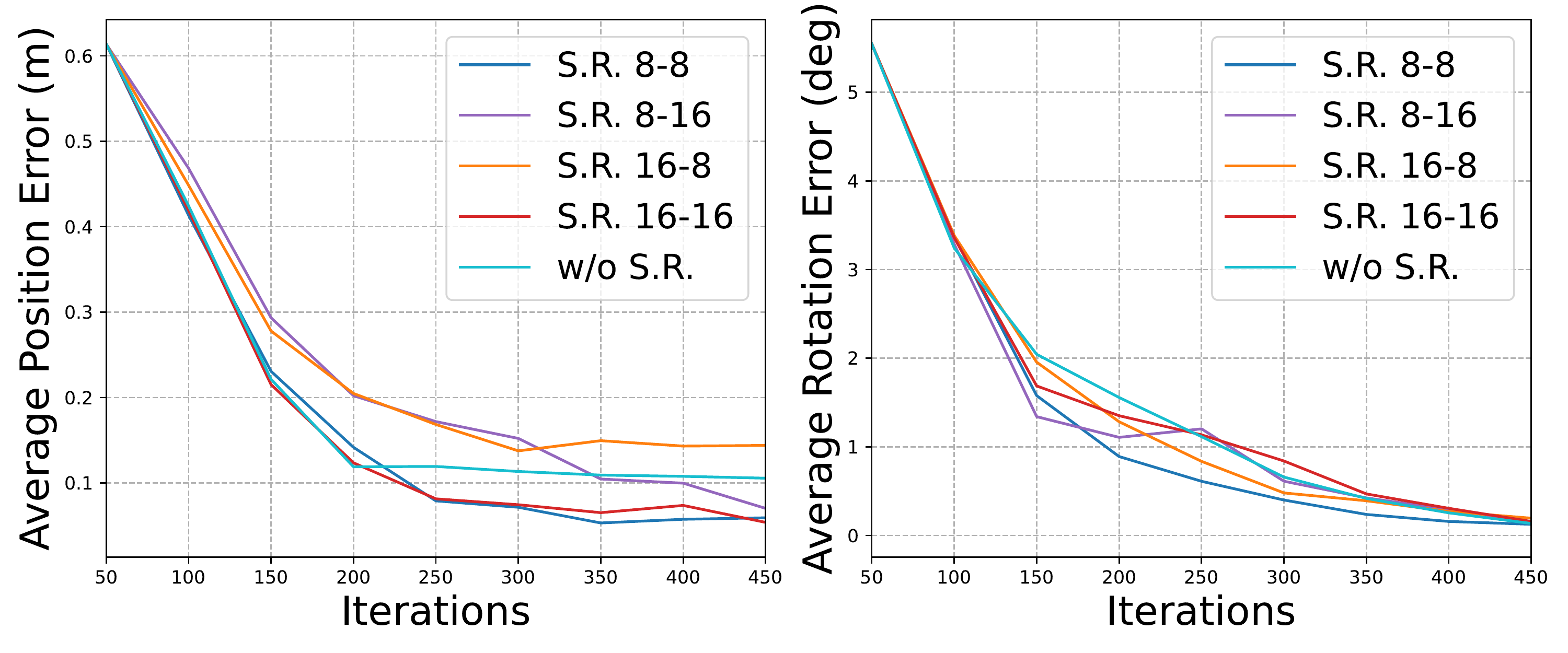}
    \caption{
    Analysis of safe-region volume rendering on pose convergence performance. S.R.8-16 denotes safe-region volume rendering with 8 ``coarse'' samples and 16 ``fine'' samples. w/o S.R denotes classical volume rendering with 64 ``coarse'' samples and 64 ``fine'' samples.
    }
    \label{fig:supp_safe_region}
\end{figure}

We analyze the impact of safe-region volume rendering on the pose convergence w.r.t. the number of sample points.
The experiment is conduct by optimizing bed poses from the iG-Synthetic bedroom with 1024 ray samples, so the classical volume rendering is also capable to optimize without out-of-memory error.
As shown in Fig.~\ref{fig:supp_safe_region}, even when only sample one fourth of the points (S.R.8-8) along each ray, we can still achieve similar convergence performance both on position error and rotation error.

\subsection{Scene Arrangement}

\begin{figure*}[htbp]
    \centering
    \includegraphics[width=0.80\linewidth, trim={0 0 0 0}, clip]{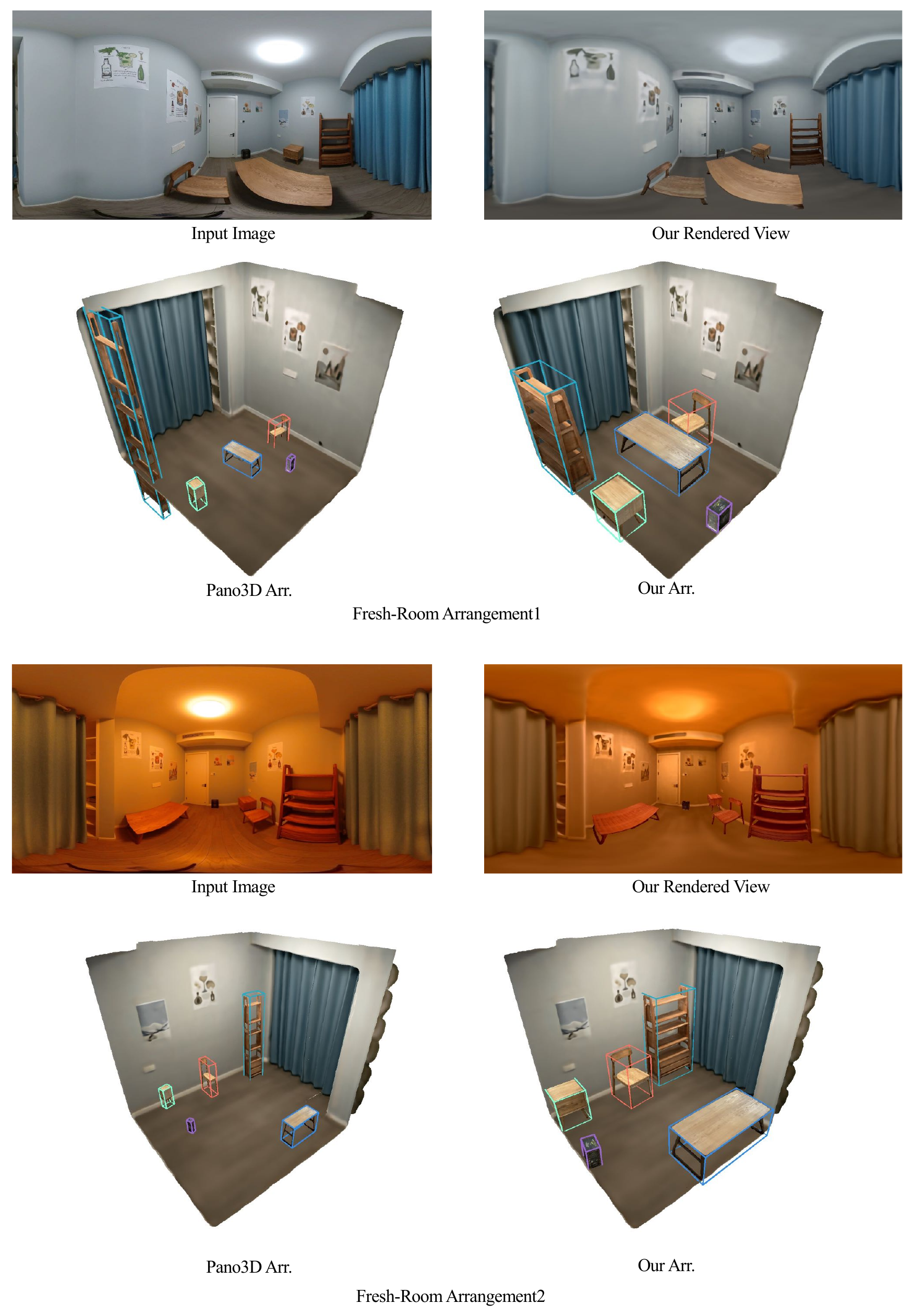}
    \caption{
    More examples of Scene Arrangement on the Fresh-Room dataset.
    }
    \label{fig:supp_scene_arr_3}
\end{figure*}

\begin{figure*}[htbp]
    \centering
    \includegraphics[width=0.80\linewidth, trim={0 0 0 0}, clip]{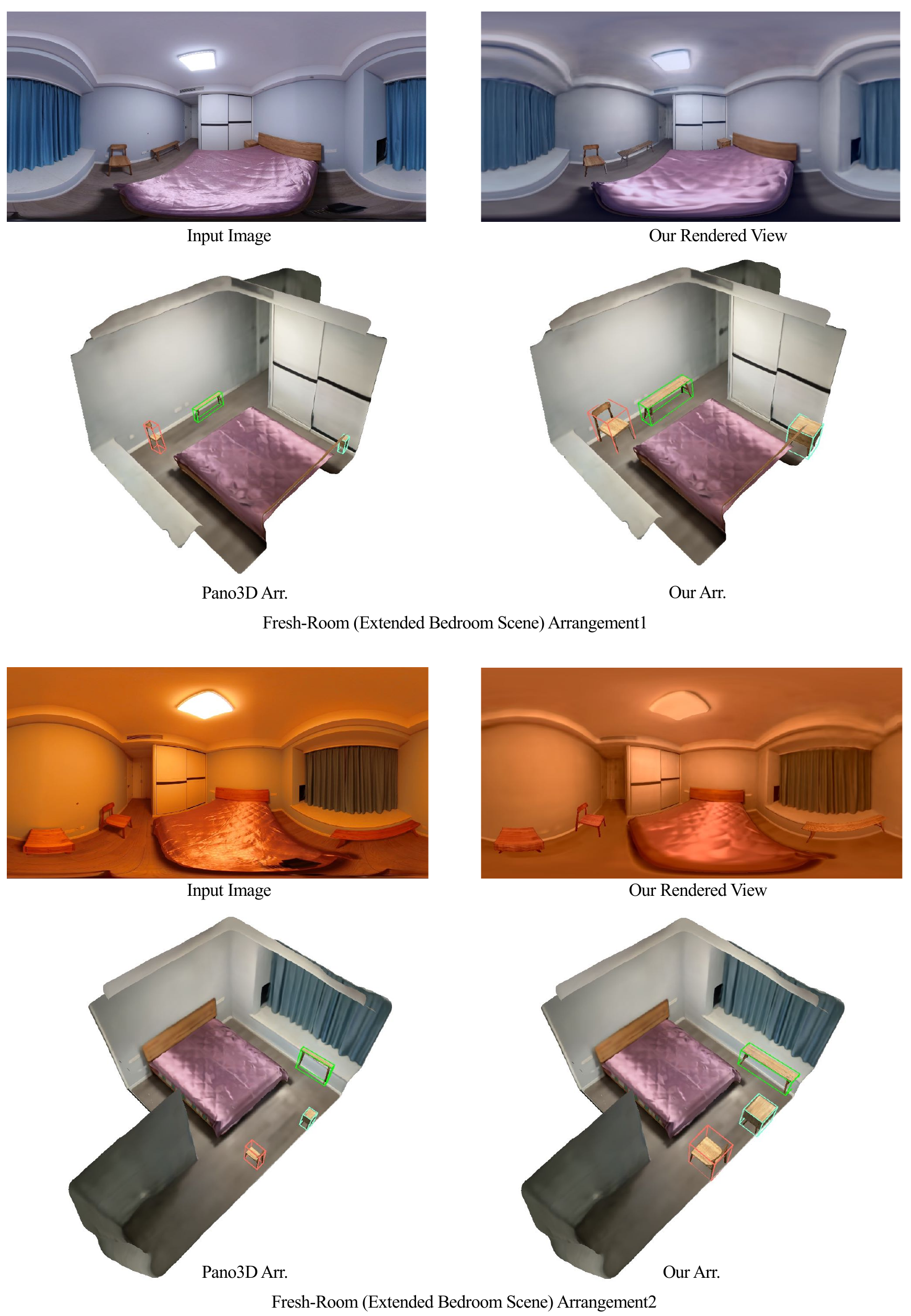}
    \caption{
    More examples of Scene Arrangement on the Fresh-Room dataset (extended bedroom scene).
    }
    \label{fig:supp_scene_arr_bedroom}
\end{figure*}

We show more qualitative comparisons with DeepPanoContext (Pano3D)~\cite{deeppanocontext} in Fig.~\ref{fig:supp_scene_arr_1}, Fig.~\ref{fig:supp_scene_arr_3}, and Fig.~\ref{fig:supp_scene_arr_bedroom} (\textit{extended bedroom scene of Fresh-Room dataset}).
It is clear that DeepPanoContext can not precisely estimate object poses (e.g., the bed and the bottom cabinet are tilted in iG-Synthetic Bedroom, see Fig.~\ref{fig:supp_scene_arr_1}), and the estimated size are distorted in the Fresh-Room dataset (see Fig.~\ref{fig:supp_scene_arr_3} and Fig.~\ref{fig:supp_scene_arr_bedroom}).
In contrast, as our neural-rendering-based holistic optimization takes both photometric and physical constraint into consideration, our method successfully estimates object arrangement with correct poses and accurate scales.

\begin{figure*}[htbp]
    \centering
    \includegraphics[width=0.85\linewidth, trim={0 0 0 0}, clip]{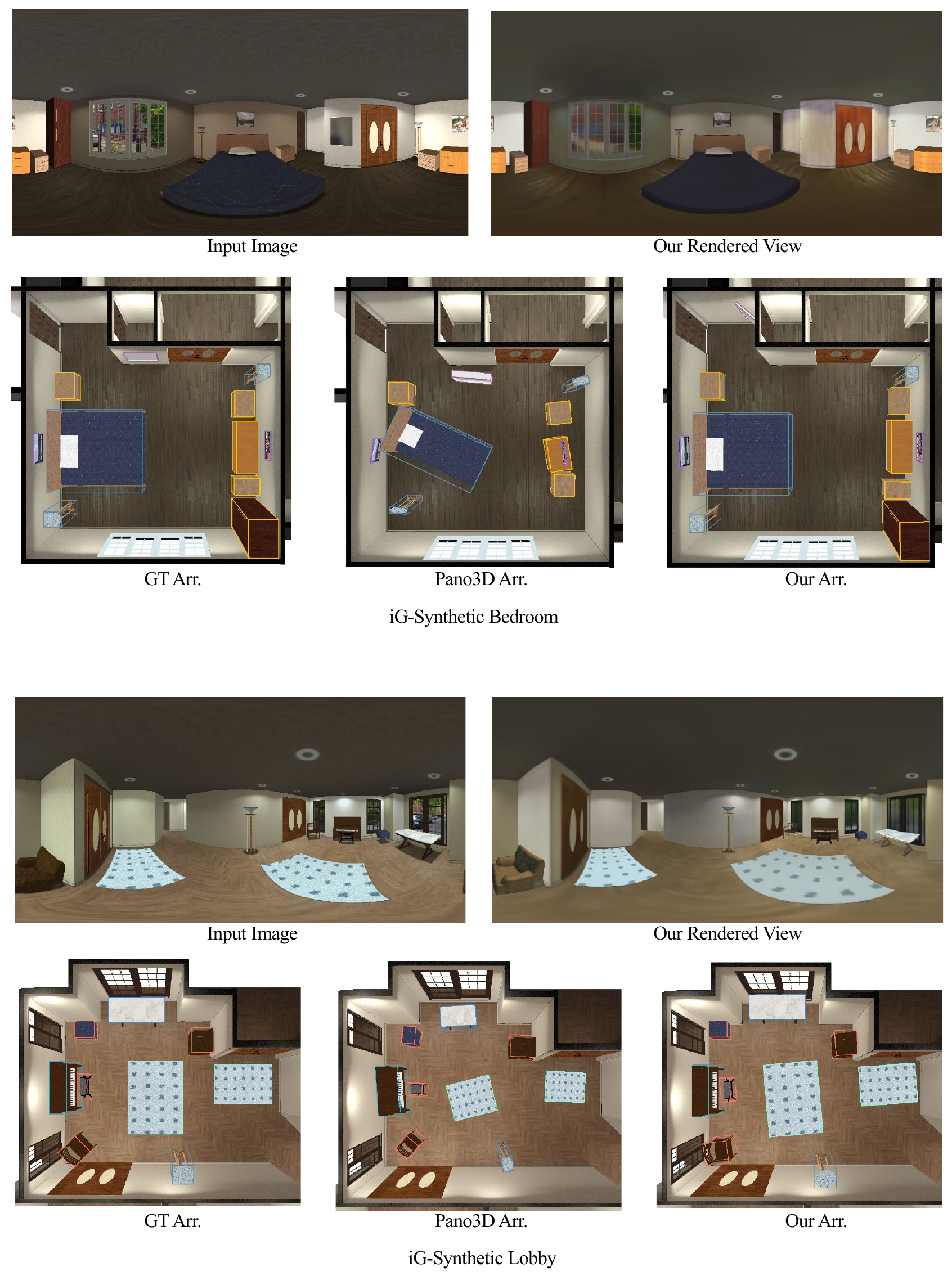}
    \caption{
    More examples of Scene Arrangement on the iG-Synthetic dataset.
    }
    \label{fig:supp_scene_arr_1}
\end{figure*}

\begin{figure*}[htbp]
    \centering
    \ContinuedFloat
    \captionsetup{format=cont}
    \includegraphics[width=0.85\linewidth, trim={0 0 0 0}, clip]{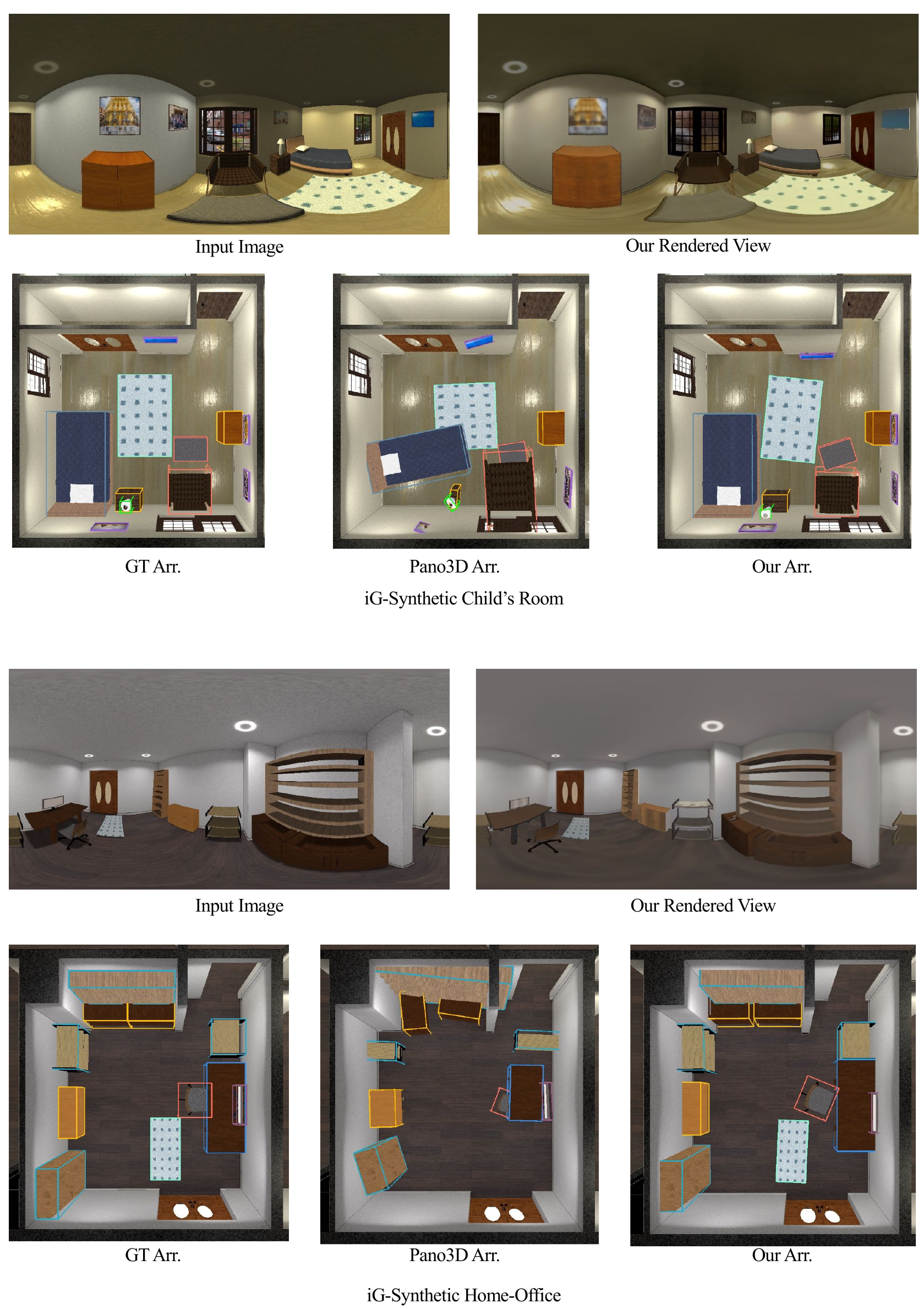}
    \vspace{-1.0em}
    \caption{
    More examples of Scene Arrangement on the iG-Synthetic dataset.
    }
    \label{fig:supp_scene_arr_2}
\end{figure*}

\subsection{More Examples on Lighting Augmentation}

\begin{figure*}[htbp]
    \centering
    \includegraphics[width=0.8\linewidth, trim={0 0 0 0}, clip]{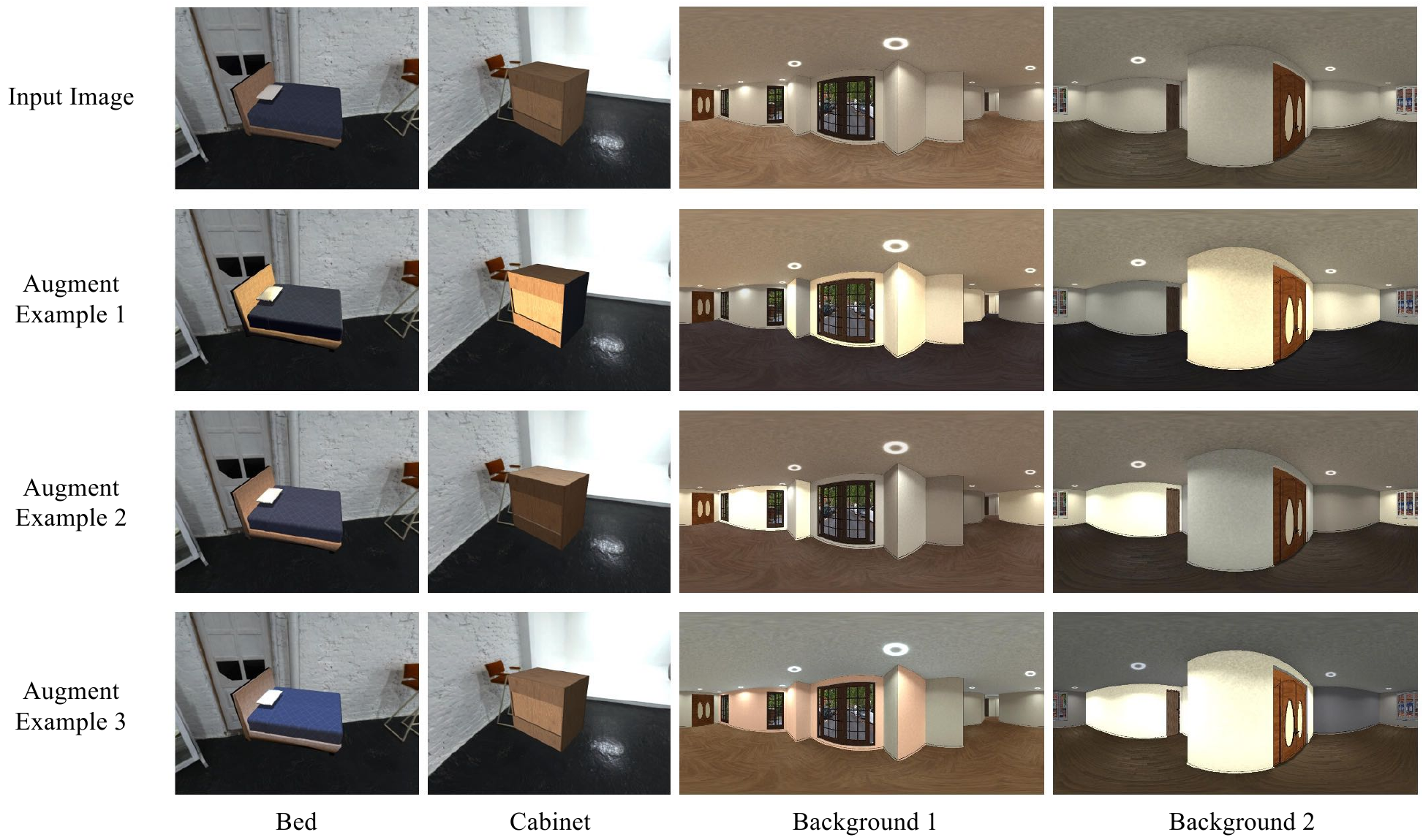}
    \caption{
    More examples of lighting augmentation on the iG-Synthetic dataset.
    }
    \label{fig:supp_light_aug_ig}
\end{figure*}

\begin{figure*}[htbp]
    \centering
    \includegraphics[width=0.8\linewidth, trim={0 0 0 0}, clip]{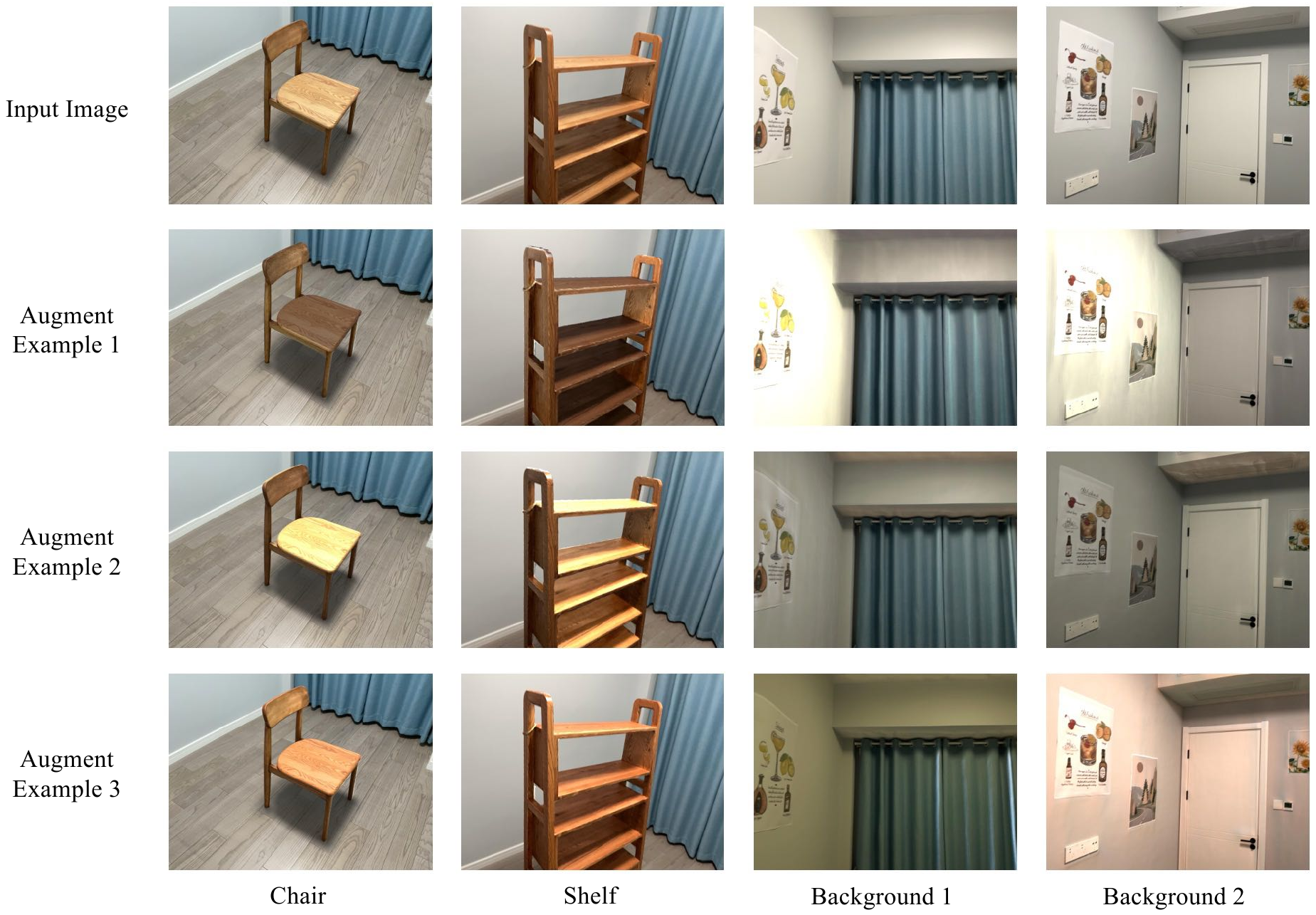}
    \caption{
    More examples of lighting augmentation on the Fresh-Room dataset.
    Note that for the 3D arrangement visualization, we use the object meshes extracted from our neural implicit renderer.
    }
    \label{fig:supp_light_aug_real}
\end{figure*}

We show more examples of lighting augmentation in Fig.~\ref{fig:supp_light_aug_ig} and Fig.~\ref{fig:supp_light_aug_real}, where the scene background and objects have been augmented with different lighting conditions.

\section{Social Impact}
Our method may suffer from potential privacy risks, since it involves recording sensitive information of the scene. To mitigate this, the user should be notified in advance and real systems should ensure that the data does not leave the users' device and could be deleted upon request.


\end{document}